\documentclass[11pt,a4paper]{article}

\usepackage{amsmath,amssymb,amsfonts}
\usepackage{booktabs}
\usepackage{multirow}
\usepackage{array}
\usepackage{tabularx}
\usepackage{graphicx}
\usepackage{geometry}
\usepackage{authblk}   % multiple-affiliation author block via \affil
\usepackage{hyperref}
\usepackage{listings}
\usepackage{xcolor}
\usepackage[T1]{fontenc}
\usepackage[utf8]{inputenc}

\hypersetup{
  colorlinks=true,
  linkcolor=blue!60!black,
  citecolor=blue!60!black,
  urlcolor=blue!60!black
}

\newcommand{\figfont}{\ttfamily\scriptsize}

\graphicspath{{./figures/}{./}}

\title{A Structural Dynamics Graph World Model: Unified Modeling, Constrained Rollout, and Interpretable Calibration}
\author{%
  Wei Wang,\, Yaosen Chen\thanks{Corresponding author: \texttt{chenyaosen@sobey.com}. All author emails share the \texttt{@sobey.com} domain: \texttt{wangwei}, \texttt{chenyaosen}, \texttt{yanghan}, \texttt{liuyuegen}, \texttt{luomingli}, \texttt{jiaoxinxin}, \texttt{wenxuming}, \texttt{liuming}.},\,%
  Han Yang,\, Yuegen Liu,\, Mingli Luo,\, Xinxin Jiao,\, Xuming Wen,\, Ming Liu%
}
\affil{Sobey Media Intelligence Laboratory}
\date{}

\begin{document}
\maketitle

\begin{abstract}
\noindent
The state evolution of a complex system typically arises jointly from the object's own laws, explicit relational propagation, domain conservation, and as-yet-unmodeled error. Forcing all sources into a single black box makes mechanism attribution and hard-constraint preservation unauditable over long-horizon rollouts; forcing every mechanism into one equation family discards mature domain solvers and heterogeneous models. 

We propose \textbf{SD-GWM}, a Structural Dynamics Graph World Model realized as an executable structural contract: nodes declare self-dynamics $S$, edges declare neighbor graph-coupled dynamics $N$---both fixed-form mechanism assets (rules, ODEs, domain solvers) calibrating only authorized parameters. An optional bounded residual $R$ concentrates learnability, while a global projection $\Pi_\Omega$ maps states back to feasibility, enforcing constraints without guaranteeing accuracy gains.

\smallskip\noindent
Built on this composition--transition--projection interface and eight frozen research questions (pre-registered), SD-GWM delivers: 
(i) \textit{Heterogeneous integration}: rules and solvers plug in natively, dividing labor cleanly with $R$; 
(ii) \textit{Semantic fidelity}: disabling $R$ preserves source semantics bit-for-bit, with four theory properties stated under explicit proof/sketch/empirical boundaries; 
(iii) \textit{Auditable governance}: stepwise traces enable counterfactual fault localization (top-1 $=1.0$) without post-hoc approximations.

\smallskip\noindent
On a semi-synthetic flood testbed and USGS streamflow, SD-GWM reduces constraint violations to floating-point tolerance in analytical tests and to zero in semi-synthetic and real-data cases. Persistence matches SD-GWM in calm periods (autocorrelation dominates), but during a 254-day extreme-flood shift persistence and all neural baselines collapse (90-min RMSE $892$--$3007$\,cfs) while SD-GWM holds at $108$\,cfs ($8$--$28\times$ gain over the collapsing baselines). The bounded residual cuts RMSE $\sim\!50\%$ only under backbone bias, confirming conditional utility. We position SD-GWM not as a universally superior forecaster, but as a verifiable substrate for auditable, constraint-safe spatiotemporal mining.


%\medskip
%\noindent\textbf{Keywords:} graph world model; executable structural contract; constrained rollout; bounded residual; semantic preservation
\end{abstract}

\section{Introduction}

World models learn a compressed spatiotemporal representation of an environment, providing a basis for prediction and planning within an internal model\cite{ref1,ref2}. Intelligence in complex systems must answer not only ``what is happening now'' but also ``what may happen next,'' and further ``what would happen if a certain action were taken, and which options satisfy the constraints and are worth executing.'' When a system is composed of objects, relations, capacities, control actions, and multiple time scales, graph structure provides a more direct relational inductive bias than a flat state vector. Interaction Networks separate object-centric from relation-centric computation\cite{ref3}; Graph Networks further provide a unified graph-computation module\cite{ref4}; and graph physics engines and graph simulators have demonstrated the use of message passing in dynamical-system identification, control, and multi-step simulation\cite{ref7,ref8}. These tools each have their strengths, yet together they leave a methodological gap. On one hand, when a graph neural network simultaneously learns its own inertia, relational propagation, constraint correction, and unknown residual, the prediction error can be optimized, but component responsibilities become difficult to audit---which mechanism bears the error, whether hard constraints are preserved, and whether the residual remains diagnosable are often unanswerable; the better a learned world model fits trajectories, the less it can explain why it is right and where it will err. On the other hand, forcing every mechanism into a single equation family discards mature domain solvers, discrete rules, and heterogeneous models---dismantling a validated stock--flow or hydraulic model into generic operators for the sake of formal unification is neither economical nor necessary. This paper seeks a path between these two extremes: neither surrendering the dynamics to a single black box nor demanding mechanism homogenization, but organizing heterogeneous models and the learnable part through an executable structural contract.

The starting point of this paper is not to propose yet another network architecture, but to organize heterogeneous, partially known, and constrained domain dynamics into an executable structural contract. Concretely, the state change of a node at each step is decomposed into three generative terms, each answering a clear question. Self-dynamics $S$ answers: if the neighbor influence is temporarily severed, how would this object evolve on its own? Graph-coupled dynamics $N$ answers: which objects can affect it through which valid relations? The minimal residual correction $R$ answers: after $S$ and $N$ generate a candidate, how should the mismatch that persists stably and is supported by data be correctably amended. $R$ is not a third world explanation---only by defining the residual after the main mechanisms can it be prevented from becoming an all-absorbing black box. There is a boundary that differs from most hybrid models: $S$ and $N$ are declarative fixed-form mechanism assets---rules, state machines, ODEs, or domain solvers---that calibrate only their authorized continuous parameters, and the learnable component is concentrated in the optional $R$; a global projection then maps the state into the feasible state space, enforcing hard constraints but not guaranteeing more accurate predictions. The formal construction of the composition, transition, and projection of the three is given in Section~3. This paper thereby turns the ``graph world model'' from a learning framework into a compilable, constrainable, and auditable operational closed loop---its learnability is confined within the declared edges and the residual budget, rather than surrendering all dynamics to a free network.

\begin{figure}[h]
\centering
\includegraphics[width=0.95\linewidth]{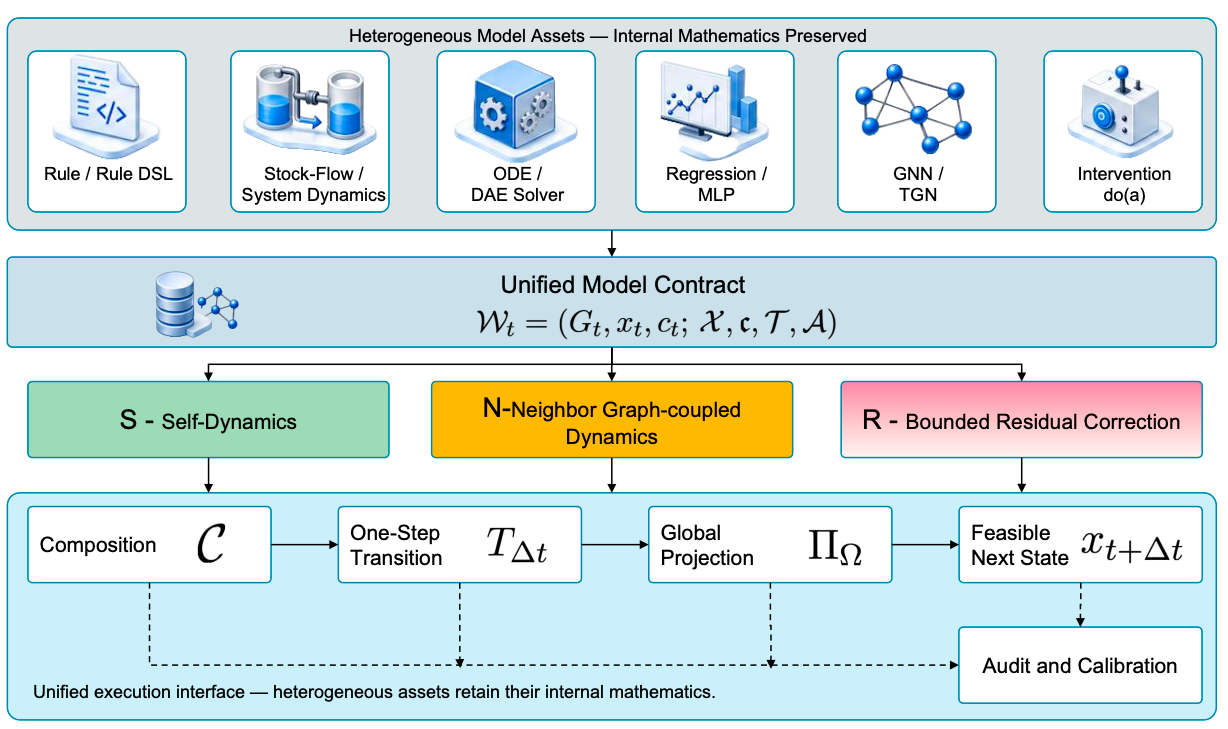}
\caption{The two-layer unified framework of SD-GWM. The upper layer retains heterogeneous model assets; the lower layer executes S/N/R composition, one-step transition, projection, audit, and calibration.}
\label{fig:unified}
\end{figure}

The contribution of this paper is to organize heterogeneous models, mechanism responsibilities, hard constraints, intervention inputs, and calibration governance into a unified operational closed loop; its novelty lies in the division of the interface and responsibility boundaries, not in a single network structure:

\begin{itemize}
\item \textbf{An executable typed structural contract.} We propose the S/N/R decomposition on dynamic property graphs together with a unified composition--transition--projection interface, allowing rules, stock--flow, ODE/domain-solver, and learning models to retain their internal mathematics and be integrated in a ``pluggable but not constraint-bypassing'' manner (the default implementations and admissible families of each operator are shown in Table~\ref{tab:op-family}). Its value: mature domain models need not be dismantled and rewritten to accommodate a unified framework, and the learnable module need not bear constraints that rules or conservation laws should own; the two divide labor at the interface rather than conflate inside a black box. The overall architecture is shown in Figure~\ref{fig:unified}.
\item \textbf{Compilation and semantic preservation.} We give the compilation from the structural contract to an executable transition operator and establish four theory properties with explicit boundaries---the exact computational embedding of a System Dynamics step, the invariant of the feasible set after correct projection, the recursive error inequality under a local Lipschitz assumption, and the single-step contribution upper bound of the residual---and distinguish four evidence levels: proof, proof sketch, empirical assumption, and engineering design principle. Its value: the conditions under which constraint preservation and residual boundedness hold are written explicitly, so the reader can judge under which assumptions a guarantee holds, rather than taking it as an unconditional universal theorem.
\item \textbf{Auditable calibration, intervention, and fault localization.} Building on the structural contract, parameters are partitioned by governance authority into frozen, calibratable, approval-required, and learnable-residual classes, with component-level execution traces retained at each step; interventions reuse the same transition model, and fault localization is based on counterfactual traces rather than black-box explanation. Its value: which mechanism a prediction primarily comes from, whether a deviation is attributable to $S$/$N$/$R$ or to the constraint projection, and along which paths an intervention propagates can all be answered quantitatively.
\end{itemize}

The three contributions are not parallel products but progressive facets of the same framework: the compilation and theory properties of Contribution~2 build on the contract interface of Contribution~1, and the audit of Contribution~3 reuses the component-level execution traces of the former two.

To make the above contributions verifiable, all model configurations and decision criteria were frozen before viewing the test results; failures or divergences must not be silently deleted, and positive and negative evidence are equally auditable---eight pre-registered research questions (RQ1--RQ8, covering compilation semantic fidelity, heterogeneous composition, constraints and interventions, structural fault localization, constrained residual, and the real-world applicability of declarative fixed mechanisms on semi-synthetic flood, public real data, and a cross-conservation-law two-domain pair; see Table~\ref{tab:protocol} in the Experiments chapter) are designed accordingly.

The experiments test the real-world applicability of declarative fixed mechanisms on two realistic scenarios: a semi-synthetic urban flood testbed (mechanism-controlled, with observations generated by deterministic noise) and USGS public real streamflow (five stations in the Potomac River basin, real third-party data). Both compare a declarative backbone calibrated from observations against neural baselines including persistence/linear/MLP/GCN/LSTM/STGCN/T-G-CN under the same protocol. The core finding is that the declarative backbone has zero constraint violation and is significantly more robust under distribution shift---on the USGS 29-day calm period, persistence ties or slightly outperforms due to high flow autocorrelation, but under the distribution shift of the 254-day period containing extreme floods, persistence and all neural baselines collapse (90-min RMSE rises to $892$--$3007$ cfs), whereas SD-GWM holds at $108$ cfs, about $8$--$28\times$ better. We emphasize that this paper claims the feasibility of a unified contract, constraint preservation, and auditable division of labor, rather than cross-scenario accuracy superiority; this result is confined to the examined small-sample semi-synthetic and single-real-watershed settings and is not extrapolated to a universal cross-scenario accuracy advantage.

The value of this framework is not presupposed to be ``always improving accuracy.'' The frozen experiments give an explicit characterization of the applicability condition of the constrained residual: when the backbone declaration has systematic bias, the bounded residual can reduce error without breaking constraints; when the budget is not binding, the bounded and unbounded residuals perform comparably. This paper therefore commits to a narrower but more verifiable conclusion: the structural contract can provide mechanism accounting, semantic preservation, and fault localization within a unified constraint and intervention interface, while predictive accuracy must still compete with baselines scenario by scenario.

\section{Related Work}

\subsection{World Models and Graph Relational Inductive Biases}

Early world models compressed visual environments into latent representations and trained policies within a model-generated internal environment\cite{ref1}; PlaNet's latent transition contained both deterministic and stochastic paths and planned actions in latent space\cite{ref2}. Such methods emphasize learning predictable states from observations. This paper does not argue with them about ``which latent variable is best,'' but studies how to organize an executable state transition when objects, relations, constraints, and domain equations are partially declarable.

The object--relation-separated Interaction Network takes the graph directly as dynamics input\cite{ref3}; the relational inductive-bias work treats entities, relations, and composition rules as the basic units of structured computation\cite{ref4}. MPNN unifies many graph neural architectures into a message--aggregation paradigm\cite{ref5}, and GCN performs convolution via local graph aggregation, with the reported implementation's computation scaling linearly with the number of edges\cite{ref6}. Learnable graph physics engines are used for system identification and trajectory optimization\cite{ref7}, and GNS represents particles as nodes and predicts physical dynamics via learned message passing\cite{ref8}. For dynamic graphs, TGN represents a dynamic graph as a timestamped event sequence and combines memory modules with graph operators\cite{ref9}. On the spatiotemporal-graph prediction side, STGCN alternates gated convolutions and graph convolutions to capture spatiotemporal dependencies\cite{ref17}, and T-G-CN combines graph convolution with gated recurrence to model spatiotemporal evolution\cite{ref19}, both common spatiotemporal-graph prediction baselines; on the non-graph sequence-modeling side, LSTM learns long-range dependencies with gated recurrent units\cite{ref18} and is also a commonly used sequence backbone in spatiotemporal prediction. This paper borrows the message--aggregation interface but confines learnable messages within declared edges and responsibility boundaries.

\subsection{Constraints, System Dynamics, and Long-Horizon Rollout}

Explicitly constrained neural dynamics can enforce constraints in Cartesian coordinates via Lagrange multipliers\cite{ref10}; neural projection methods enforce physics constraints learned from data by iteratively correcting predicted positions\cite{ref16}. These works support the background that ``constraints can enter model execution,'' but do not provide a general accuracy guarantee for the business-state projection of this paper. This paper sets the execution of declared constraints and the accuracy of predicting real systems as two separate evaluation dimensions. Closer to this paper's $R$ is the ``physics prior plus learning residual'' line: physics-informed neural networks constrain the solution space by entering the PDE residual into the loss\cite{ref20}, and Lagrangian/Hamiltonian neural networks take symmetry conservation as a differentiable prior with an unbounded residual part\cite{ref21}. Their common feature is that the prior and the residual are both embedded in the same differentiable computational graph, and the residual has no explicit budget. This paper takes the opposite boundary: $S$ and $N$ are declarative fixed-form mechanism assets that do not go through gradients, and the learnable component is concentrated in $R$, which is bounded dimension-wise by $B_\varepsilon$---``how much the residual may change at most'' is fixed by a design parameter rather than implicitly determined by the training loss. This makes mechanism responsibilities separately auditable, at the cost that the forms of $S$ and $N$ must be explicitly declared from domain knowledge and cannot be partially inferred from data as in physics-informed networks.

System Dynamics studies system behavior through feedback, flow graphs, delays, and policy experiments\cite{ref11}. This paper absorbs stock--flow, feedback, delay, and policy-experiment semantics but does not set a global ODE or stock--flow as the sole computational master. System Dynamics is a computationally embeddable and preserved model; this paper does not replace System Dynamics, nor does it claim to cover its participatory modeling, organizational learning, and policy negotiation traditions.

Autoregressive rollout encounters inconsistency between training history and generation history. Scheduled sampling progressively replaces real history with model-generated history to reduce the train/inference input-distribution gap\cite{ref12}. This paper's goal of maintaining state legality in multi-step rollout is also a long-horizon rollout problem, but the mechanism differs. Existing graph world models support both unstructured and graph-structured states and express task actions via action nodes\cite{ref13}. A 2026 survey preprint proposes organizing graph world models by spatial, physical, and logical-relational inductive biases and by connector, simulator, and reasoner roles\cite{ref14}; this is only the authors' proposed taxonomy, not an established consensus, and the accessibility of that preprint should be judged by its formally published version. Another 2026 preprint decomposes the long-horizon error of fixed-edge graph world models into a topology-related factor and a model spectral-norm factor, and discusses node--edge error feedback and rollout-consistency regularization in dynamic edges\cite{ref15}; again, the citation stands as a preprint. This paper draws on the idea of attributing error to structural and model factors, but in Proposition~3 it explicitly distinguishes the topology proxy, the spectral-norm proxy of learned weights, and the formal Lipschitz constant---the latter two are not interchangeable---and this distinction is a theoretical choice of this paper, not directly implied by that preprint's decomposition.

\subsection{Interpretability and Auditability}

The ``audit'' goal of this paper has a clear boundary with post-hoc explanation methods. Graph neural network explanation methods such as GNNExplainer search, on a trained model, for the subgraph and features most important to a prediction\cite{ref22}, and counterfactual explanations seek ``what input change would alter the decision'' without modifying the model's internals\cite{ref23}. Their common feature is that explanation occurs after prediction and does not change the model's execution structure---the model remains a black box, and the explanation is a local characterization of its input--output relation. This paper takes the opposite path: auditability is not a post-hoc addition but is built in by the structural contract at design time---the contributions of $S$, $N$, and $R$ are recorded at the component level at each step, and fault localization is based on executing a repair on each candidate and comparing the error reduction (counterfactual tracing), rather than on post-hoc attribution of a black-box output. The cost is that the mechanism forms must be explicitly declared, so a model that is itself a free network cannot be audited; the benefit is that attribution comes directly from the execution structure and does not depend on the post-hoc approximation of explanation methods.

Across the above work, graph world model research has accumulated tools in representation learning, message passing, and long-horizon error analysis, but still lacks an interface that unifies heterogeneous mechanisms, hard constraints, interventions, and calibration governance into an auditable operational closed loop: existing methods either surrender all dynamics to a free network and lose responsibility accounting, or constrain the solution space with differentiable physics priors but with no explicit residual budget, or demand mechanism homogenization and lose domain models. The structural contract of this paper targets exactly this gap.

\section{Problem Definition and Preliminaries}

Before presenting the method, we first fix the dynamic-world representation, the form of nodes and edges, and the constrained one-step transition interface that runs through the paper; the decomposition, composition, projection, and calibration of subsequent chapters all unfold on this representation.

\textbf{Definition 1 (Dynamic-world snapshot).} At time $t$, the world consists of the graph instance and state instance at that time, together with the spatial, temporal structure and asset set that are invariant throughout:
\begin{equation}
\mathcal{W}_t=(G_t,x_t,c_t;\,\mathcal X,\mathfrak{c},\mathcal T,\mathcal A),\qquad
G_t=(\mathcal{V}_t,\mathcal{E}_t),\quad \mathcal{E}_t\subseteq \mathcal{V}_t\times \mathcal{V}_t\times\Lambda .
\tag{1}
\end{equation}

Before the semicolon are the instances that vary with time: the graph $G_t$, the state $x_t\in\mathcal X$, and the context $c_t\in\mathfrak{c}$; after the semicolon are the definitional structures invariant throughout: the state space $\mathcal X$, the context space $\mathfrak{c}$, the temporal structure $\mathcal T$ (the discrete step and timestamp grid that determine the cadence of transitions), and the asset set $\mathcal A$ (the formal assets carrying mechanisms, including rules, state machines, ODEs, solver code, and parameters $\Theta=\Theta_S\cup\Theta_N\cup\Theta_R$). $\Lambda$ is the set of relation types, the value domain of declared edges. A forward prediction takes $\mathcal{W}_t$ as a read-only snapshot; structural modifications to nodes, edges, and rules occur in the calibration governance process after prediction. Dynamics nodes include state nodes, as well as intervention nodes that, once overridden by an intervention, are treated as state; input, modulator, and evidence enter as exogenous input, gating input, and calibration observation, respectively, and are not mixed into the updated state---this separation keeps exogenous driving and endogenous evolution apart at the interface rather than conflating them in the state vector (the responsibilities of the five node types are shown in Table~\ref{tab:node-types}).

\textbf{Definition 2 (Node state and typed edge).} For a dynamics node $v$, the state and the typed edge from $u$ to $v$ are written
\begin{equation}
x_v(t)\in\mathcal X_v\subseteq\mathbb R^{d_v}\times\mathcal D_v,
\qquad e=(u,v,\tau_e),\quad \tau_e\in\Lambda .
\tag{2}
\end{equation}

Continuous and discrete states may coexist: $\mathbb R^{d_v}$ is the continuous component and $\mathcal D_v$ the discrete component (e.g., on/off, gear, mode label), both juxtaposed in the same node state. The per-node space $\prod_v\mathcal X_v$ only guarantees local legality of each node; the global feasible state space $\Omega\subseteq\prod_v\mathcal X_v$ additionally contains cross-node hard constraints such as conservation, shared capacity, and safety interlocking---local legality does not imply global legality, and it is exactly this gap that makes the projection $\Pi_\Omega$ necessary. The main symbols and their responsibilities are shown in Table~\ref{tab:symbols}.

\begin{table}[h]
\centering
\caption{Main symbols and responsibilities.}
\label{tab:symbols}
\begin{tabularx}{\linewidth}{@{}l l >{\raggedright\arraybackslash}X@{}}
\toprule
Symbol & Meaning & Boundary \\
\midrule
$S,N,R$ & Self / neighbor graph-coupled / residual dynamics & Non-overlapping responsibilities \\
$f_S,f_N,f_R$ & Function implementations of $S,N,R$ & Composed via $\mathcal{C}$ \\
$B_\varepsilon$ & Dimension-wise soft-clipping operator & Definition~8, $|B_\varepsilon(z)_i|\le\varepsilon_i$ \\
$\mathcal{C}$ & Composition operator & Default weighted addition; may be modulation/switching/solving \\
$\mathcal{F}$ & Dynamics field & Total field composed by $\mathcal{C}$ (Definition~4) \\
$T_{\Delta t}$ & One-step transition & May invoke rules, discrete models, or solvers \\
$\Pi_\Omega$ & Global projection & Legalization, not mechanism explanation \\
$\Omega$ & Feasible state space & Per-node constraints plus cross-node hard constraints \\
$\mathcal{D}_v$ & Discrete-state component of node $v$ & Juxtaposed with $\mathbb R^{d_v}$ in $x_v$ \\
$\mathcal{T}$ & Temporal structure & Discrete step and timestamp grid, invariant throughout \\
$\mathcal{A}$ & Asset set & Mechanism formal assets and parameters $\Theta$ \\
$\mathfrak{c}$ & Context space & Instance $c_t\in\mathfrak{c}$ \\
$\Lambda$ & Relation-type set & Value domain of declared edges \\
\bottomrule
\end{tabularx}
\end{table}

\begin{table}[h]
\centering
\caption{The five node types and their responsibilities. state and intervention enter the updated state; input/modulator/evidence are not mixed into the updated state, separating exogenous driving from endogenous evolution at the interface.}
\label{tab:node-types}
\begin{tabular}{lll}
\toprule
Node type & Responsibility & Enters updated state \\
\midrule
state & Endogenous evolving state & Yes \\
intervention & Treated as state after intervention override & Yes (after override) \\
input & Exogenous input & No \\
modulator & Gating input & No \\
evidence & Calibration observation & No \\
\bottomrule
\end{tabular}
\end{table}

\subsection{Unified Prediction Interface}

Definition~1 packages the world as a snapshot $\mathcal{W}_t$; prediction then extracts the state $x_t$, the graph $G_t$, and the context $c_t$ from it as the input to a one-step transition. The core of the unified prediction interface is the constrained one-step transition, where $\mathcal{F}$ is the dynamics field given by Definition~4, i.e., the total field composed from the three generative terms $S$, $N$, and $R$ via the composition operator $\mathcal{C}$; it first appears here as a placeholder interface, and its construction is given in Definition~4. This paper uses $\mathcal{F}$ without a subscript to denote an operator invariant across time, and $\mathcal{F}_t$ with a subscript to denote its value at time $t$.

\textbf{Definition 3 (Constrained one-step transition).} Given a snapshot and a step size, the unified prediction is
\begin{equation}
x_{t+\Delta t}
=
\Pi_{\Omega_{t+\Delta t}}
T_{\Delta t}(x_t,G_t,c_t;\mathcal{F}).
\tag{3}
\end{equation}

This form does not require all components to be rewritten as a single ODE: for a rule model, $T_{\Delta t}$ executes a rule transition; for a discretized domain model, it invokes an existing one-step mapping; for a strongly coupled composite node, it invokes a preserved domain solver. The $(x_t,G_t,c_t)$ before the semicolon is the time input, and the $\mathcal{F}$ after the semicolon is the configuration---the same $\mathcal{F}$ can serve transitions at different times, which is precisely the point of ``unified interface, heterogeneous implementation.''

\section{Structural Dynamics Graph World Model}

Section~3 fixed only the world representation and the one-step transition interface; it did not yet describe how the dynamics field inside $T_{\Delta t}$ is composed from the various sources. This section decomposes the one-step state change into self-dynamics, neighbor graph-coupled dynamics, and bounded residual correction, gives their composition, message passing, conservation, recursion, and intervention forms, and delimits which parameters are calibratable and which belong to structural decisions requiring approval.

\subsection{S/N/R Decomposition and Composition}

This subsection gives the formal decomposition of the three generative terms and the composition operator; it is the basis for the subsequent message passing, conservation, and residual constructions.

\textbf{Definition 4 (Structural dynamics decomposition).} The total dynamics field is composed from self-dynamics, neighbor graph-coupled dynamics, and bounded residual correction via the composition operator $\mathcal{C}$; the default is the weighted addition $\mathcal{C}_w$. Let $f_S,f_N,f_R$ denote the function implementations of $S,N,R$, and $B_\varepsilon$ the dimension-wise soft-clipping operator given by Definition~8 (the residual must be bounded by it before being added); the state $x_t$, graph $G_t$, trajectory buffer $\mathcal B_t$, context $c_t$, and parameters $\Theta_*$ are passed in as common inputs after the semicolon (where $G_t$ is read only by $f_N,f_R$, and $\Theta_*$ is the full parameter set of Definition~11, $\Theta=\Theta_{\mathrm{frozen}}\cup\Theta_{\mathrm{learn}}$):
\begin{equation}
\mathcal F_t=\mathcal{C}\big(f_S,\,f_N,\,B_\varepsilon f_R;\,x_t,G_t,\mathcal B_t,c_t,\Theta_*\big),
\qquad
\mathcal{C}_{w}=\alpha_S f_S+\alpha_N f_N+\alpha_R B_\varepsilon f_R.
\tag{4}
\end{equation}

Here $\mathcal{C}$ is the general signature of the composition operator and $\mathcal{C}_w$ is its default implementation (weighted addition); other implementations (modulation/switching/solving) are given below and in Table~\ref{tab:op-family}. $\mathcal B_t$ is the short-term trajectory buffer needed to satisfy the maximum lag. $f_S$ describes the holding, decay, mean reversion, state machine, or domain-local transition of a node after its in-edges are severed; $f_N$ describes the messages, flows, modulation, triggering, or structural switching that occur only along valid edges. It must be emphasized that the functional forms of $S$ and $N$ are declared from domain knowledge as fixed mechanisms---rules, state machines, ODEs, or domain solvers---and this paper does not train $S$ or $N$ themselves as free neural networks; the learnable component is concentrated in the calibration of authorized continuous parameters and the optional residual $R$. $R$ only compensates the error that still persists stably and is data-supported after the two, and its learning interface is reserved as a replaceable operator, but replacement requires approval and must not change the bounded and projection-finishing constraints. $\alpha_S,\alpha_N,\alpha_R$ are the relative composition weights of the three generative terms, continuous parameters at the composition layer that reduce to ordinary addition when defaulted to 1 and may enter empirical-risk-minimization calibration (see Definition~11). Beyond weighted addition, $\mathcal{C}$ may also be parameter modulation (the neighbor term modulates the self-dynamics parameters rather than entering the dynamics field directly), mode switching (selecting a set of composition sub-operators by context), or domain-solving composition (replacing explicit weighting with an algebraic/conservation solve); switching the composition mode itself (weighted $\to$ modulation $\to$ switching $\to$ solving) is a discrete structural decision that changes the form of $\mathcal{C}$, requires expert approval, and is not part of continuous calibration.

\subsection{Typed Messages, Gating, Delays, and Aggregation}

After the decomposition gives the three generative terms, we must specify how $f_N$ passes messages along edges and aggregates them at nodes.

\textbf{Definition 5 (Edge message).} For a valid edge of type $\lambda$, the message of rule $r$ is
\begin{equation}
m_{u\rightarrow v}^{r}(t)
=
\gamma_r(x_t,c_t)
\phi_r\!\left(
x_u(t-\ell_r),x_v(t),c_t;\theta_r
\right),
\qquad 0\le\gamma_r\le1 .
\tag{5}
\end{equation}

$\ell_r$ determines which historical time is read and $\phi_r$ determines how it is transformed; the two must not be conflated. The historical read $x_u(t-\ell_r)$ is provided by the trajectory buffer $\mathcal B_t$ of Definition~4, and the length of $\mathcal B_t$ is determined by the graph-wide maximum lag $\max_r\ell_r$. When the gating is zero the message is off, and continuous gating allows context modulation. When the initial history is insufficient, a pre-declared padding strategy must be used, not one selected by error at test time.

\textbf{Definition 6 (Neighbor aggregation).} Multiple rules on the same edge are first composed according to pre-declared semantics, and then the target node aggregates its in-edges:
\begin{equation}
M_v(t)=\operatorname{Agg}_v
\left(\left\{
\operatorname{RuleAgg}_{uv}(\{m_{u\rightarrow v}^{r}(t)\}_r)
\right\}_{u\in\operatorname{In}(v)};\eta_v\right).
\tag{6}
\end{equation}

Aggregation may be a weighted sum, a capacity-limited sum, a maximum-risk, or a constrained learnable aggregation; $\eta_v$ is the configuration parameter of the aggregation (e.g., the weights of a weighted sum or a capacity bound). Locality comes from the input set rather than post-hoc explanation: unconnected nodes cannot produce a direct single-step message. Attention or GNNs may implement the message and aggregation, but their inputs remain constrained by the graph and roles.

\subsection{Stock--Flow Conservation and Composite Models}

Not all edges carry ordinary messages: when a node is a stock and an edge is a flow, total conservation must additionally be required.

\textbf{Definition 7 (Local conservation at a stock node).} Only when the state is a stock and the edge is a flow do we require
\begin{equation}
\frac{dx_v}{dt}
=
\sum_{e:u\rightarrow v}q_e
-\sum_{e:v\rightarrow w}q_e
+u_v,\qquad
\frac{d\sum_vx_v}{dt}=\sum_vu_v .
\tag{7}
\end{equation}

\begin{figure}[h]
\centering
\includegraphics[width=0.95\linewidth]{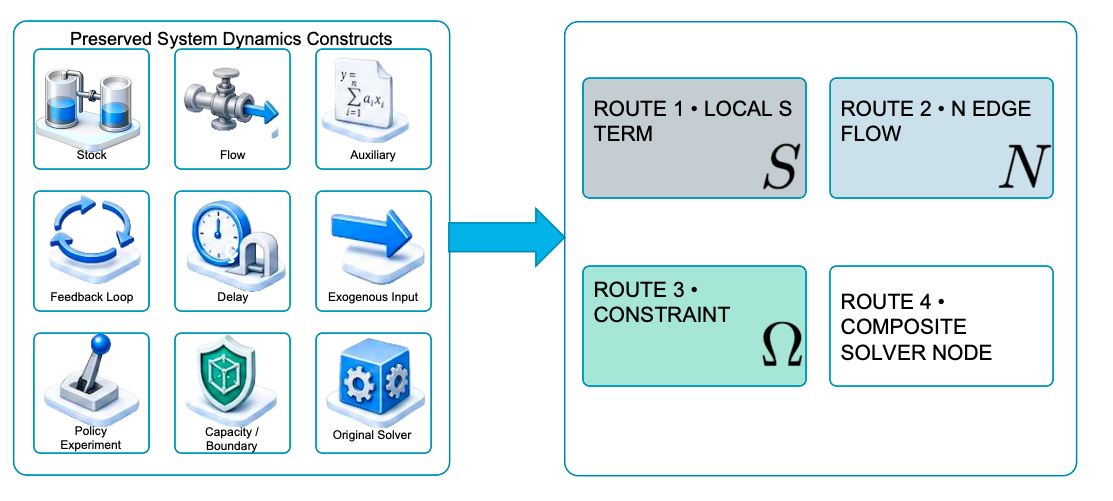}
\caption{The computational mapping of System Dynamics constructs into $S$, $N$, $\Omega$, or composite solver nodes; the original model's mathematics and solver are preserved.}
\label{fig:sd-embedding}
\end{figure}

Control commands, risk messages, and modulation relations are not required to conserve. Equation~(7) is written in continuous time to carry on the System Dynamics tradition, but in this paper's discrete one-step transition $T_{\Delta t}$ it is executed in a discretized form according to an established numerical scheme (e.g., explicit Euler or a built-in solver scheme); the continuous form of the conservation law defines ``what should be conserved,'' and the discretization together with the projection $\Pi_\Omega$ ensures ``it is indeed conserved after execution.'' Mature stock--flow, ODE, SWMM, or HEC-RAS\footnote{SWMM (Storm Water Management Model) is the U.S. Environmental Protection Agency (EPA) urban stormwater-runoff management model; HEC-RAS (Hydrologic Engineering Center's River Analysis System) is the U.S. Army Corps of Engineers Hydrologic Engineering Center's river hydraulic analysis system. Both are standard numerical models in hydrology and hydraulics, cited here as examples of mature domain models with dedicated numerical procedures.} models that contain strong algebraic loops, DAEs, or dedicated numerical procedures should be retained as composite nodes with their internal solvers preserved, not dismantled for the sake of formal unification. The mapping from System Dynamics constructs to the entry points of structural dynamics is shown in Figure~\ref{fig:sd-embedding}.

\subsection{Bounded Residual, Recursive Rollout, and Intervention}

Conservation handles the known mechanisms; the residual handles the stable mismatch after the mechanisms. This subsection gives the bounded form of the residual, the multi-step recursion, and the intervention transformation.

\textbf{Definition 8 (Bounded residual).} The raw residual output $z=f_R(\cdot)$ passes through dimension-wise soft clipping:
\begin{equation}
B_\varepsilon(z)
=
\varepsilon\odot\tanh(z\oslash\varepsilon),
\qquad |B_\varepsilon(z)_i|\le\varepsilon_i .
\tag{8}
\end{equation}

Bounded does not mean the residual is a correct mechanism. $f_R$ is implemented in this paper's experiments by the default graph-message-passing residual model, and its learning interface is reserved as a replaceable operator, but replacement requires approval and must not change the bounded and projection-finishing constraints; RQ5 also uses only this default implementation. If the residual is persistently same-signed, topologically clustered, or persistently hitting its ceiling, one should first check the $S$ parameters, missing edges, omitted context, and observation issues; structural modification must go through Diagnose--Validate--Approve, not an automatic expansion of the residual capacity.

\textbf{Definition 9 (Recursive rollout).} Given a pre-declared future context path, the $k$-step prediction is
\begin{equation}
\hat{x}_{t+(k+1)\Delta t}
=
\Pi_{\Omega_{t+(k+1)\Delta t}}
T_{\Delta t}\!\left(
\hat{x}_{t+k\Delta t},G_{t+k\Delta t},c_{t+k\Delta t};\mathcal{F}
\right),
\quad \hat{x}_t=x_t .
\tag{9}
\end{equation}

The $\mathcal{F}$ after the semicolon is consistent with Definition~3: recursive rollout reuses the same dynamics field, only replacing the observed state at each step with the model's own previous-step output $\hat{x}_{t+k\Delta t}$. This equation states that long-horizon prediction feeds the previous-step model output as the next-step input; it does not automatically guarantee error convergence. Training may use generated-history curricula or rollout-consistency regularization, but these are training designs that reduce distribution shift, not stability proofs.

\textbf{Definition 10 (Intervention transformation).} An intervention $a$ is a declared transformation of the graph, context, or parameters before prediction:
\begin{equation}
\mathcal I_a:
(G_t,c_{t:t+H},\Theta)
\mapsto
(G_t^a,c_{t:t+H}^a,\Theta^a),
\qquad
\Delta_a=\hat{x}_{t+H}^{a}-\hat{x}_{t+H}^{\varnothing}.
\tag{10}
\end{equation}

Edge breaking, pump switching, capacity expansion, and parameter override all reuse the same model. $\Delta_a$ may obtain a causal interpretation only when the edge direction has causal semantics, key confounders are controlled, and the transformation corresponds to an executable action; the flood results of this paper claim only conditional what-if rollout.

\subsection{Calibration and Parameter Responsibilities}

The preceding decomposition defines which parameters enter the dynamics field; this subsection delimits which parameters can be calibrated and which belong to structural decisions requiring approval.

\textbf{Definition 11 (Frozen--learnable partition).} Domain-trusted parameters are subject to declared numerical bounds and variability constraints: continuous calibration is performed only within the learnable subset $\Theta_{\mathrm{learn}}$ and must not cross the bounds; values beyond the bounds must fall back to an expert-approved structural decision rather than being automatically relaxed by the optimizer.
\begin{equation}
\hat{\Theta}_{\mathrm{learn}}
=
\arg\min_{\Theta_{\mathrm{learn}}\in\mathcal K}
\frac1n\sum_{i=1}^{n}
\mathcal L\!\left(
\hat{x}_{i,t+\Delta t};
y_{i,t+\Delta t}
\right)
+\lambda\mathcal R(\Theta_{\mathrm{learn}}),
\quad
\Theta=\Theta_{\mathrm{frozen}}\cup\Theta_{\mathrm{learn}} .
\tag{11}
\end{equation}

The learnable terms include the composition weights $\alpha_S,\alpha_N,\alpha_R$ (the default weighted addition of Eq.~(4), defaulting to 1 reduces to ordinary addition), continuous weights, delays, thresholds, aggregation parameters, and restricted residual weights; node addition/removal, edge direction, rule type, gating logic, and the switching of the composition mode (weighted $\to$ modulation $\to$ switching $\to$ solving, changing the form of $\mathcal{C}$) belong to expert-approved discrete structural decisions. Structural masking, parameter-responsibility isolation, staged freezing, minimal-residual regularization, and ablation jointly provide operational identifiability (i.e., reducing responsibility conflation through masking, freezing, ablation, and audit), but this operational identifiability is not statistical identifiability. Calibration is separated by objective into two non-joint paths: finite-difference least-squares calibration of continuous parameters adjusts only structural parameters, and empirical-risk minimization trains only the residual model; the two are not jointly optimized within the same loop. The relation between the above discrete structural decisions and the admissible families is shown in Table~\ref{tab:op-family}.
\begin{table}[!htbp]
\centering
\caption{Default implementations and admissible families of the main operators. The admissible family only lists implementation types and does not specify particular algorithms; ``default implementation'' refers to the implementation used in this paper's experiments and is non-exclusive; an ``alternative implementation'' must satisfy the corresponding admissibility conditions and be approved, and must not change the composition order or the projection finishing.}
\label{tab:op-family}
\begin{tabularx}{\linewidth}{@{}l l >{\raggedright\arraybackslash}X >{\raggedright\arraybackslash}X@{}}
\toprule
Operator & Symbol & Default implementation & Admissible family \\
\midrule
Self-dynamics & $f_S$ & Decay / mean reversion & Holding, state machine, domain-local transition \\
Graph-coupled backbone & $f_N$ & Weighted message aggregation & Capacity-limited sum, attention aggregation, conservative flow \\
Residual model & $f_R$ & Graph-message-passing residual & Other bounded learnable corrections (must pair with $B_\varepsilon$) \\
Composition operator & $\mathcal{C}$ & Weighted addition $\mathcal{C}_w$ & Modulation, switching, domain solving \\
One-step transition & $T_{\Delta t}$ & Rule one-step transition & Discretized domain mapping, composite-node solver \\
Global projection & $\Pi_\Omega$ & Interval clipping & Enumerative alignment, invariant recovery, safety gate \\
\bottomrule
\end{tabularx}
\end{table}

\section{Theory Properties}

The previous chapter defined the engineering constructions of decomposition, composition, projection, and calibration, but did not yet state what they guarantee mathematically and what they do not. This chapter gives four theory properties with explicit boundaries (stated as propositions) and explicitly distinguishes proof, proof sketch, empirical assumption, and engineering design principle, so that the reader can judge under which conditions each guarantee holds.

\subsection{Projection and Computational Embedding}

This subsection gives the formal definition of the projection and proves that, with the residual disabled, structural dynamics can exactly embed System Dynamics.

\textbf{Definition 12 (Euclidean projection).} For a non-empty closed convex feasible set $\Omega$,
\begin{equation}
\Pi_\Omega(z)=\arg\min_{x\in\Omega}\|x-z\|_2 .
\tag{12}
\end{equation}
Here $z$ is any point to be projected (e.g., a candidate state that may leave the feasible set after a one-step transition), $x$ is a candidate within the feasible set, and the projection returns the feasible point $x$ closest to $z$.

Non-convex safety switching or enumerative correction may also serve as engineering legalization operators, but cannot directly inherit the non-expansiveness property of closed-convex projection. The user-side projection type is restricted to declarative strategies (interval clipping, enumerative alignment, invariant recovery, safety gate, etc.); solving-type or custom projection operators are reserved system extensions that may be enabled only after approval and run after the built-in projection, not replacing the global legalization finishing step.

\textbf{Proposition 1 (Computational embedding of System Dynamics under residual-off and solver matching).}

\noindent\textbf{Assumption:} The original stock--flow/System Dynamics model has a deterministic one-step mapping $F_{\mathrm{SD}}^{\Delta t}$ given the initial state, exogenous input, parameters, step size, numerical solver, and update order; the structural-dynamics composite node uses the same mapping, the same numerical precision, and enables no additional residual or altered projection (the exogenous input corresponds to the exogenous component of this paper's context $c_t$). Then there exists a structural-dynamics instance satisfying
\begin{equation}
T_{\Delta t}^{\mathrm{SD}}(x_t,c_t;\Theta_{\mathrm{SD}})
=F_{\mathrm{SD}}^{\Delta t}(x_t,c_t;\Theta_{\mathrm{SD}}),
\qquad \hat{x}_{t+k\Delta t}=x_{t+k\Delta t}^{\mathrm{SD}} .
\tag{13}
\end{equation}

\noindent\textbf{Conclusion:} Under the above degenerate configuration, there exists a structural-dynamics instance that is numerically identical to the original System Dynamics in both the one-step mapping and the multi-step trajectory.

\noindent\textbf{Proof.} Register $F_{\mathrm{SD}}^{\Delta t}$ as the one-step transition of the composite node, set $R$ to zero, and keep the projection as the legalization operation already present in the original model. The base step $k=1$ holds by the interface being identical; if the step-$k$ state is identical, then the same state, input, parameters, and solver order produce the identical step-$(k+1)$. Induction completes the argument. This proposition is a computational-semantic embedding and does not claim to replace the methodological tradition of System Dynamics.\hfill$\square$

\textbf{Proposition 2 (Invariant of the feasible set after correct projection).}

\noindent\textbf{Assumption:} At each step $\Pi_{\Omega_{t+\Delta t}}$ returns an element of $\Omega_{t+\Delta t}$ for any candidate state, and the initial state is feasible. Then
\begin{equation}
\hat{x}_{t+k\Delta t}\in\Omega_{t+k\Delta t},
\qquad k=1,2,\ldots .
\tag{14}
\end{equation}

\noindent\textbf{Conclusion:} As long as the projection is correctly implemented and the initial value is feasible, the predicted state lies in the feasible set at any step and is not carried out of it by the illegality of the transition candidate.

\noindent\textbf{Proof.} The first step holds by the range definition of the projection; assuming step $k$ is feasible, the next step, regardless of whether the transition candidate is illegal, still enters the next-time feasible set after correct projection. Induction completes the argument. This proposition guarantees the execution of declared constraints, not that the candidate model is accurate, and certainly not that the real-world constraints are declared completely.\hfill$\square$

\subsection{Conditional Error Recursion and Residual Upper Bound}

Unless otherwise stated, the vector norm in this section and in all error bounds in the paper is the absolutely monotone $\ell_2$ norm, i.e., when each component's absolute value is held down the norm does not increase ($|x_i|\le|y_i|\,\forall i\Rightarrow\|x\|\le\|y\|$); the corresponding operator gains refer to the local Lipschitz gains, on a specified trajectory segment, induced by this norm, not global constants.

\textbf{Proposition 3 (Conditional recursive error inequality).}

\noindent\textbf{Assumption:} The reference truth trajectory lies stepwise within the corresponding feasible set, so the projection does not move the truth (the projection only pulls candidates outside the feasible set back and does not alter points already inside), i.e., $\Pi_{\Omega_{t+k\Delta t}}(x_{t+k\Delta t})=x_{t+k\Delta t}$, which provides an anchor for the error analysis that does not vary with the projection; the reference one-step mapping and the model one-step mapping have local Lipschitz upper bounds $L_k$ in the neighborhood of the examined trajectory; the projection is $K_{\Pi,k}$-Lipschitz in that neighborhood; and the combined upper bound of the non-residual model mismatch, graph-structure error, and exogenous-path error at step $k$ is $\delta_k$. Define $\varepsilon_k^R$ as the pre-projection one-step difference, at step $k$ on the same reference input, graph, and context, produced by enabling the residual relative to disabling it. Let $e_k=\|\hat{x}_{t+k\Delta t}-x_{t+k\Delta t}\|$; then
\begin{equation}
e_{k+1}
\le
K_{\Pi,k}\big(L_k e_k+\delta_k+\|\varepsilon_k^{R}\|\big),
\quad
e_H\le
\left(\prod_{q=0}^{H-1}K_{\Pi,q}L_q\right)e_0
+
\sum_{j=0}^{H-1}
\left(\prod_{q=j+1}^{H-1}K_{\Pi,q}L_q\right)
K_{\Pi,j}(\delta_j+\|\varepsilon_j^R\|).
\tag{15}
\end{equation}

\noindent\textbf{Conclusion:} Under the local Lipschitz and other conditions, the multi-step prediction error is upper-bounded by a product amplification term of the initial error and a weighted sum of the per-step errors. Note that what is given here is an error upper bound, not a convergence proof: $L_k$ may still exceed one, and the error may keep amplifying with the number of steps, so no universal convergence is implied.

\noindent\textbf{Proof sketch.} Insert the reference truth, fixed by the same projection, both before and after the projection; apply the triangle inequality, the local Lipschitz assumption, and the projection Lipschitz upper bound to obtain the one-step recursion; unfolding it step by step yields the product term of the initial error and the weighted sum of the per-step errors, where the empty product is taken as one (in the summation term of Eq.~(15), when $j=H-1$, $\prod_{q=j+1}^{H-1}$ is empty and defined as 1). Under the same-initial-value condition $\hat{x}_t=x_t$ of Definition~9, $e_0=0$, and Eq.~(15) simplifies to a summation containing only the per-step errors. If $\Omega$ is a non-empty closed convex set and the Euclidean projection is used, the non-expansive upper bound of the projection may be taken; but $L_k$ may still exceed one, so no universal convergence conclusion is in the equation. The projections in this paper's experiments (RQ1--RQ8) are all interval clipping, i.e., the Euclidean projection on the box constraint $[\ell,u]$, which is a non-empty closed convex case, so the $K_{\Pi,k}$-Lipschitz assumption of Proposition~3 holds under the experimental configuration and $K_{\Pi,k}\le1$; non-convex projections (enumerative alignment, safety gate, etc.) are retained as engineering extensions, are not enabled in this paper's experiments, and under them this assumption requires separate verification. The graph-adjacency spectral quantity and the learned-weight spectral norm can only serve as empirical proxies for $L_k$; the spectral-norm proxy of this paper is not a formal Lipschitz constant.

\textbf{Proposition 4 (Single-step contribution of the bounded residual).}

\noindent\textbf{Assumption:} The dimension-wise bounded operator of Eq.~(8) is used with each $\varepsilon_i>0$, where $\varepsilon$ is a pre-declared design parameter not automatically adjusted by single-step calibration. Fix the same non-residual inputs and construct two trajectories, residual-on and residual-off; $a_R$ is defined as the local Lipschitz gain of the composition operator between the residual terms of the two trajectories, $L_R$ as the local Lipschitz gain of the one-step transition on the line segment connecting the corresponding points of the two trajectories, and $K_\Pi$ as the local Lipschitz gain of the projection on the line segment connecting the corresponding points of the two candidate-state trajectories. Then, under the above absolutely monotone $\ell_2$ norm, the single-step output perturbation caused by the residual satisfies
\begin{equation}
\|\Delta x_{t+\Delta t}^{R}\|
\le K_\Pi L_R a_R\|\varepsilon\| .
\tag{16}
\end{equation}

\noindent\textbf{Conclusion:} The single-step output perturbation caused by the residual is upper-bounded by $\|\varepsilon\|$ and the product of the three local gains of composition, transition, and projection---``how much the residual may change at most'' is fixed by the design parameter $\varepsilon$.

\noindent\textbf{Proof.} Equation~(8) gives $\|B_\varepsilon f_R\|\le\|\varepsilon\|$; applying the local-gain upper bounds of composition, transition, and projection in turn completes the argument. Here $a_R$ covers the dimensional mapping from the residual output to the state quantity: the raw residual output $z=f_R(\cdot)$, after being clipped by $B_\varepsilon$, has dimensions consistent with $\varepsilon$; when it is added to the candidate state by the composition operator $\mathcal{C}$, if a dimensional conversion exists (e.g., the unit of $\alpha_R$ in the weighted addition), that conversion is folded into $a_R$, so both sides of Eq.~(16) have the same dimensions. If the composition is solving-type and $R$ is disabled, the term is zero. Under the default weighted addition (Eq.~(4)), $a_R$ is the absolute value of the composition weight $|\alpha_R|$, and the bound reduces to the trivial estimate in the linear case; the value of the proposition is to constrain the maximum perturbation amplitude of the residual under modulation-type, switching-type, and solving-type composition. This bound only constrains the contribution amplitude; it does not prove the residual direction is correct, nor does it imply OOD robustness.\hfill$\square$

The preceding definitions and propositions decompose the one-step rollout into several operators, each allowing the selection of one implementation within its operator family rather than fixing a single formula (the global form of a few operators is not open to alternatives; see the note after Table~\ref{tab:op-family}). The framework fixes the composition order of the operators and the admissibility conditions of each, leaving the choice of implementation open: each operator has a default implementation and may be replaced by an alternative satisfying the same admissibility conditions---i.e., the generalization of the replaceable operator $f_R$ of Definition~8 and the reserved projection extension of Definition~12. Table~\ref{tab:op-family} gives the default implementations and admissible families of the main operators as the culmination of the above definitions and propositions.

The above replaceability is the formal locus of the ``pluggable but not constraint-bypassing'' principle: an alternative implementation must not change the order by which $f_S,f_N,B_\varepsilon f_R$ are composed via $\mathcal{C}$, transitioned via $T_{\Delta t}$, and finished via $\Pi_\Omega$; the residual must be bounded by $B_\varepsilon$ before being added; and the output of any alternative operator must still enter the unified $\mathcal{C}$ composition and the $\Pi_\Omega$ constraint-preserving finish. The finishing roles of $f_S$ and $\Pi_\Omega$ are not open to alternatives in this paper; the form switching of $\mathcal{C}$ (weighted $\to$ modulation $\to$ switching $\to$ solving) is an approval-type discrete structural decision, not continuous calibration. The alternative implementations of the remaining operators must declare the admissibility conditions they satisfy, such as boundedness ($f_R$), output domain ($f_S,\Pi_\Omega$), and idempotence ($\Pi_\Omega$). This paper's experiments are completed on the default implementations of each operator and do not enable alternative implementations. Mature System Dynamics models need not be dismantled to accommodate the above interface; their preserved mapping to the entry points of structural dynamics is shown in Table~\ref{tab:sd-mapping}.

\begin{table}[h]
\centering
\caption{Preserved mapping between System Dynamics and structural dynamics.}
\label{tab:sd-mapping}
\begin{tabularx}{\linewidth}{@{}l l X@{}}
\toprule
System Dynamics construct & Structural-dynamics entry & Preserved content \\
\midrule
Stock & state node or composite state & Stock semantics and units \\
Flow & conservative-type $N$ message & Source decrement, target increment \\
Auxiliary & input/modulator & Exogenous or intermediate quantity \\
Feedback loop & directed cycle & Feedback direction \\
Delay & $\ell_r$ and buffer & Delay semantics \\
Policy experiment & intervention transformation & Conditional scenario comparison \\
Solver & $T_{\Delta t}$ or composite node & Numerical algorithm and calibration parameters \\
\bottomrule
\end{tabularx}
\end{table}

\section{Experiments}

The theory propositions of Section~5 give only conditional guarantees; whether they can be realized in engineering, and to what degree, must be tested experimentally. This chapter designs research questions around the claims of the structural contract: whether compilation preserves source-model semantics, whether heterogeneous mechanisms compose correctly, whether constraints and interventions work end-to-end, whether structural representation improves diagnostic capability, when the constrained residual is beneficial, and whether declarative fixed mechanisms can compete with neural baselines on a semi-synthetic high-fidelity case and on public real data. All model configurations and decision criteria were frozen before viewing the test results; failures or divergences must not be silently deleted, and positive and negative evidence are equally auditable. RQ1--RQ4 are tested one by one on mechanism-controlled synthetic or analytical environments, whose advantage is that the mechanism truth is available and can be compared exactly; RQ5 tests the applicability condition of the constrained residual; RQ6 tests real-world applicability on a semi-synthetic flood testbed; RQ7 tests real-world applicability on public real data. The statistical units and evidence levels are shown in Table~\ref{tab:protocol}, and the research questions and decision criteria in Table~\ref{tab:hypotheses}.

\begin{table}[h]
\centering
\small
\caption{Research questions and decision criteria. RQ1--RQ4 correspond to the theory propositions of Section~5 or their degenerate cases; RQ5 is the applicability-condition test of the constrained residual; RQ6 is the real-world-applicability test of declarative fixed mechanisms on a semi-synthetic high-fidelity case; RQ7 is the real-world-applicability test of declarative fixed mechanisms on public real data.}
\label{tab:hypotheses}
\begin{tabularx}{\linewidth}{@{}l l >{\raggedright\arraybackslash}X >{\raggedright\arraybackslash}X@{}}
\toprule
ID & Type & Claim 1& Decision criterion \\
\midrule
RQ1 & Empirical test & Compilation preserves source-model semantics & Error at floating-point tolerance supports the implementation \\
RQ2 & Empirical test & Heterogeneous mechanisms compose correctly & Trajectory consistency, zero conservation and port-type violations \\
RQ3 & Empirical test & Constraints and interventions work end-to-end & Deployed violation and counterfactual error decrease in lockstep \\
RQ4 & Empirical test & Structural representation improves diagnostic capability & Trace diagnosis outperforms residual magnitude and random baselines \\
RQ5 & Applicability condition & When the constrained residual is beneficial & When the backbone has systematic bias, the residual reduces error without breaking constraints \\
RQ6 & Real-world applicability & Declarative fixed mechanisms compete with neural baselines & The observation-calibrated declarative backbone RMSE is no worse than baselines and constraint violation is zero \\
RQ7 & Real-world applicability & Declarative fixed mechanisms work on real data & On real data, the calibrated declarative backbone multi-step RMSE is no worse than baselines and non-negativity violation is zero \\
RQ8 & Real-world applicability & Declarative mechanisms generalize across conservation laws & The same declared operator family is no worse than baselines in multi-step prediction under two conservation laws and each has zero $\Omega$ violation \\
\bottomrule
\end{tabularx}
\end{table}

\begin{table}[h]
\centering
\small
\caption{RQ1--RQ8 experimental protocol. ``Evidence type'' distinguishes synthetic/analytical/semi-synthetic/real evidence; ``independent unit'' is the statistical unit.}
\label{tab:protocol}
\begin{tabularx}{\linewidth}{@{}l >{\raggedright\arraybackslash}X >{\raggedright\arraybackslash}X >{\raggedright\arraybackslash}X >{\raggedright\arraybackslash}X@{}}
\toprule
Research question & Goal & Evidence type & Independent unit & Main output \\
\midrule
RQ1 & Compilation semantic fidelity & Synthetic or analytical & Deterministic state coordinate & Max absolute error \\
RQ2 & Heterogeneous composition & Synthetic & Deterministic transition step & Trajectory RMSE, conservation and port violations \\
RQ3 & Constraints and interventions & Analytical & Intervention scenario & Deployed violation, counterfactual error, projection norm \\
RQ4 & Structural fault localization & Synthetic & Seed-level diagnosis aggregation & top-1/top-3 accuracy, macro F1 \\
RQ5 & Constrained residual & Synthetic & Seed--regime rollout & RMSE, budget usage, constraint violation \\
RQ6 & Declarative-mechanism flood & Semi-synthetic & Test event & RMSE, constraint violation, counterfactual direction \\
RQ7 & Declarative-mechanism real data & Real & Test time block & RMSE, non-negativity violation rate \\
RQ8 & Declarative-mechanism cross-conservation-law & Real & Test time block & RMSE, conservation violation rate \\
\bottomrule
\end{tabularx}
\end{table}

RQ1--RQ3 are deterministic synthetic or analytical evidence, each running twice under the same configuration with bit-for-bit agreement; RQ4 and RQ5 are repeated on five random seeds, with seed-level aggregation as the statistical unit. The fault severity of RQ4 is sampled with the seed, so that the deterministic diagnosis method faces faults of different severities, producing genuine cross-seed variance; the residual-model training of RQ5 uses a fixed random seed to ensure reproducibility. RQ6 is conducted on a mechanism-controlled semi-synthetic flood testbed, with twelve test events as the independent statistical unit, neural baselines repeated on five seeds and averaged within a seed before entering the paired test; RQ7 is the same, with twelve test time blocks as the independent unit. The statistical meaning of the two-level aggregation should be clarified: the independent unit is the test event/time block (12 in total), the five seeds are used to estimate seed variance, and the paired Wilcoxon test is paired over the 12 events/blocks (each unit's value being the within-seed average), so the effective sample size is 12, not 60. All result values come from frozen result-summary artifacts and can be independently reproduced by public scripts (the artifact list and verification are in Appendix~\ref{app:repro}).

\subsection{RQ1: Does Compilation Preserve Source-Model Semantics?}

RQ1 tests the degenerate case of Proposition~1: after decomposing the source model's update into the structural contract's $S$ (node self-mechanism) and $N$ (edge action) and setting $R$ to zero, are the compiled model and the source reference implementation numerically consistent in both the one-step mapping and the multi-step trajectory? Three source models are covered: a threshold state machine (discrete rules), a coupled dual-tank (stock--flow conservation), and a spring--mass--damper system (coupled ODE). Because the compiled side and the reference side execute the same set of floating-point operations, the same numerical precision and update order at each step, the two trajectories are bit-for-bit identical rather than approximate. The result verifies that the structural contract can exactly embed the source model under the degenerate configuration of residual-off and bit-for-bit solver matching; it does not verify a broader model family, nor does it cover the embedding behavior after the residual is enabled. The ``number of independent units'' is the number of deterministic state coordinates compared bit-for-bit on the multi-step trajectory for each source model (state machine 8, tanks 80, spring-mass 160). The maximum absolute trajectory difference is shown in Table~\ref{tab:rq1}.

\begin{table}[h]
\centering
\caption{RQ1 compilation semantic fidelity: maximum absolute trajectory difference between the compiled model and the source reference implementation. The differences are at floating-point tolerance, arising because both sides use the same solver, the same numerical precision and update order, matching bit-for-bit rather than approximately.}
\label{tab:rq1}
\begin{tabular}{lcr}
\toprule
Source model & Number of independent units & Max absolute error \\
\midrule
State machine & 8 & 0.000000 \\
Coupled tanks & 80 & 0.000000 \\
Spring-mass & 160 & $4.44\times10^{-16}$ \\
\bottomrule
\end{tabular}
\end{table}

The maximum absolute errors of all three source models are at floating-point zero or machine-precision magnitude, supporting RQ1: the compilation of the structural contract can exactly preserve the source-model semantics when the residual is disabled. This conclusion is limited to the bit-for-bit matching degenerate configuration and is not extrapolated to the residual-on case or a broader model family.

\subsection{RQ2: Can Heterogeneous Mechanisms Compose Correctly?}

RQ2 tests the core value of the structural contract relative to a single domain solver---whether heterogeneous mechanisms can compose safely within the same graph. On a single graph, rule nodes (state machines), ODE nodes (spring--mass), and stock--flow nodes (tank conservation) are mixed, with gating, delay, and external-input edges, verifying type and unit checks, scheduling and causal order, port compatibility, conservation, and numerical stability after composition. The composed model is compared against independent reference implementations for consistency on the multi-step trajectory, with 12 deterministic transition steps as the independent unit. The results are shown in Table~\ref{tab:rq2}.

\begin{table}[h]
\centering
\caption{RQ2 heterogeneous composition: maximum trajectory RMSE, maximum mass-conservation error, maximum event-order mismatch, and maximum port-type violation of the composed model relative to the independent reference implementation.}
\label{tab:rq2}
\begin{tabular}{lcr}
\toprule
Metric & Number of independent units & Maximum \\
\midrule
Max trajectory RMSE & 12 & $1.28\times10^{-16}$ \\
Max mass-conservation error & 12 & $6.02\times10^{-16}$ \\
Max event-order mismatch & 12 & 0 \\
Max port-type violation & 12 & 0 \\
\bottomrule
\end{tabular}
\end{table}

The trajectory RMSE and conservation error of heterogeneous composition are both at floating-point tolerance, with zero event-order and port-type violations, supporting RQ2: mechanism nodes and edges of different types can compose correctly within the same structural contract, and the causal order, port matching, and conserved quantities are all preserved after composition.

\subsection{RQ3: Do Constraints and Interventions Work End-to-End?}

RQ3 tests the end-to-end effectiveness of Proposition~2: in a full rollout, can the global projection enforce both single-node boundaries and cross-node conservation, and evaluate candidate violation and deployed state separately? Four projection strategies are compared: no projection, per-node clipping (clipping only single-node boundaries, not handling cross-node conservation), sequential heuristic projection, and joint convex projection. Interventions are measured by counterfactual error: for supported structural interventions, the prediction difference after intervention versus no intervention is compared with the reference difference. The results are shown in Table~\ref{tab:rq3}.

\begin{table}[h]
\centering
\caption{RQ3 constraints and interventions: maximum deployed violation, average counterfactual error, and average projection norm under four projection strategies.}
\label{tab:rq3}
\begin{tabularx}{\linewidth}{@{}>{\raggedright\arraybackslash}X c r r@{}}
\toprule
Projection strategy & Max deployed violation & Avg counterfactual error & Avg projection norm \\
\midrule
No projection & 8.000000 & 2.984173 & 0.000000 \\
Per-node clipping & 2.000000 & 1.604266 & 1.750000 \\
Sequential heuristic & 0.666667 & 0.117851 & 2.913456 \\
Joint convex projection & $\sim10^{-14}$ & $\sim10^{-14}$ & 2.984173 \\
\bottomrule
\end{tabularx}
\end{table}

The joint convex projection reduces both the deployed-state violation and the counterfactual error to floating-point tolerance in lockstep, whereas no projection leaves a violation of $8.0$, per-node clipping leaves $2.0$, and the sequential heuristic leaves $0.67$. This supports RQ3: the global projection executes the declared legality and conservation in an end-to-end rollout, and constraint and intervention consistency holds. It should be emphasized that the projection norm and the deployed violation are inversely related: the joint convex projection has the lowest violation but the highest projection norm ($2.98$), because it pulls the candidate state thoroughly back into the feasible set with the largest modification; no projection has the highest violation but a zero norm, because it does not modify the candidate at all. The norm reflects ``how much the projection changed'' rather than ``how accurate the result is''---it guarantees legality but not accuracy, and this is the clear boundary between Proposition~2 and prediction accuracy.

\subsection{RQ4: Does Structural Representation Improve Diagnostic Capability?}

RQ4 tests the auditability claim of the structural contract: whether execution-trace-based counterfactual diagnosis can localize the fault source and outperform structure-free baselines. Six classes of faults are injected into the heterogeneous graph (wrong node parameters, missing edge, wrong gating, wrong delay, sensor-mapping error, unmodeled dynamics), with fault severity sampled by seed to produce genuine cross-seed variance. Three diagnosis methods are compared: random ranking, residual magnitude (ranking by the per-candidate-coordinate residual mean), and trace counterfactual (executing a repair on each candidate and comparing the error reduction after repair). The results are shown in Table~\ref{tab:rq4}.

\begin{table}[h]
\centering
\caption{RQ4 structural fault localization: top-1 accuracy, top-3 accuracy, macro F1, false-positive rate, and post-calibration recovery of three diagnosis methods on five seeds.}
\label{tab:rq4}
\begin{tabular}{lccccc}
\toprule
Method & top-1 & top-3 & macro F1 & FPR & Recovery \\
\midrule
Random ranking & 0.100 & 0.567 & 0.088 & 1.000 & 0.094 \\
Residual magnitude & 0.333 & 1.000 & 0.158 & 1.000 & 0.320 \\
Trace counterfactual & \textbf{1.000} & \textbf{1.000} & \textbf{1.000} & \textbf{0.000} & \textbf{0.956} \\
\bottomrule
\end{tabular}
\end{table}

Trace counterfactual significantly outperforms the two baselines on top-1 accuracy, macro F1, and recovery, with zero false positives; the top-1 of random ranking is only $0.10$ and that of residual magnitude only $0.33$. This supports RQ4: the component-level execution trace of the structural contract makes the fault source localizable and outperforms the structure-free residual-magnitude and random baselines. Three points should be honestly noted. First, the top-1 accuracy of the three methods (aggregated by fault instance, excluding the fault-free clean class) is constant across the five seeds: trace counterfactual is always $1.0$, residual magnitude always $0.333$, and random ranking always $0.10$, with zero cross-seed variance---because this metric already averages over the six fault-instance classes within a seed, and the cross-seed-sampled fault-severity differences are smoothed out by the aggregation. Second, the top-1 accuracy of trace counterfactual is always $1.0$ because the counterfactual repair is insensitive to fault severity: the correct repair always reduces the error the most. Third, the cross-seed variance is not in top-1 but in the recovery magnitude: the recovery magnitude by fault class ranges from $0.88$ to $0.97$, and after seed-level aggregation it is constant at $0.956$, with its variance coming from the repairability of different fault classes.

\subsection{RQ5: When Is the Constrained Residual Beneficial?}

RQ5 tests the empirical side of Proposition~4: when the backbone declaration has systematic bias, can the budget-constrained residual $R$ reduce the error without breaking the mechanism semantics and constraints? Taking a spring--mass--damper system as the truth, the backbone deliberately declares a wrong spring parameter (spring-force edge weight $-1.5$, truth $-2.0$, a $25\%$ systematic bias), so that SD-GWM (w/o R) is persistently too soft and the error accumulates with the number of steps. The residual $R$ is trained to target the difference of the truth minus the crippled backbone (missing spring term plus damping term plus cubic term), subject to a per-dimension budget of $\varepsilon=0.05$. Four configurations of SD-GWM are compared: w/o R (residual disabled), w/ R, global (bounded global residual), w/ R, local (bounded local residual), and w/ R, unbounded (unbounded global residual), evaluated on four out-of-distribution regimes: iid (same distribution as training), missing-physics (the truth's cubic nonlinearity is strengthened, which the backbone lacks), forcing-shift (the exogenous forcing amplitude is increased), and topology-shift (the spring coefficient is changed, simulating structural mismatch); the specific parameters of each regime are in Appendix~\ref{app:hyperparameters}. The RMSE is shown in Table~\ref{tab:rq5}.

\begin{table}[h]
\centering
\caption{RQ5 constrained residual: RMSE of four SD-GWM configurations on four regimes (mean of five seeds). SD-GWM (w/ R, global) is lower than SD-GWM (w/o R) on all regimes.}
\label{tab:rq5}
\begin{tabular}{lcccc}
\toprule
Configuration & iid & Missing-physics & Forcing-shift & Topology-shift \\
\midrule
SD-GWM (w/o R) & 0.1453 & 0.2166 & 0.2360 & 0.4333 \\
SD-GWM (w/ R, global) & \textbf{0.0725} & \textbf{0.1275} & \textbf{0.1783} & \textbf{0.2834} \\
SD-GWM (w/ R, local) & 0.1835 & 0.2646 & 0.2437 & 0.4572 \\
SD-GWM (w/ R, unbounded) & 0.0725 & 0.1275 & 0.1783 & 0.2836 \\
\bottomrule
\end{tabular}
\\[2pt]
\footnotesize Note: w/o R disables the bounded residual; w/ R enables the residual, with the suffix global/local denoting the residual scope, and unbounded being the no-budget ablation.
\end{table}

\begin{figure}[h]
\centering
\includegraphics[width=0.95\linewidth]{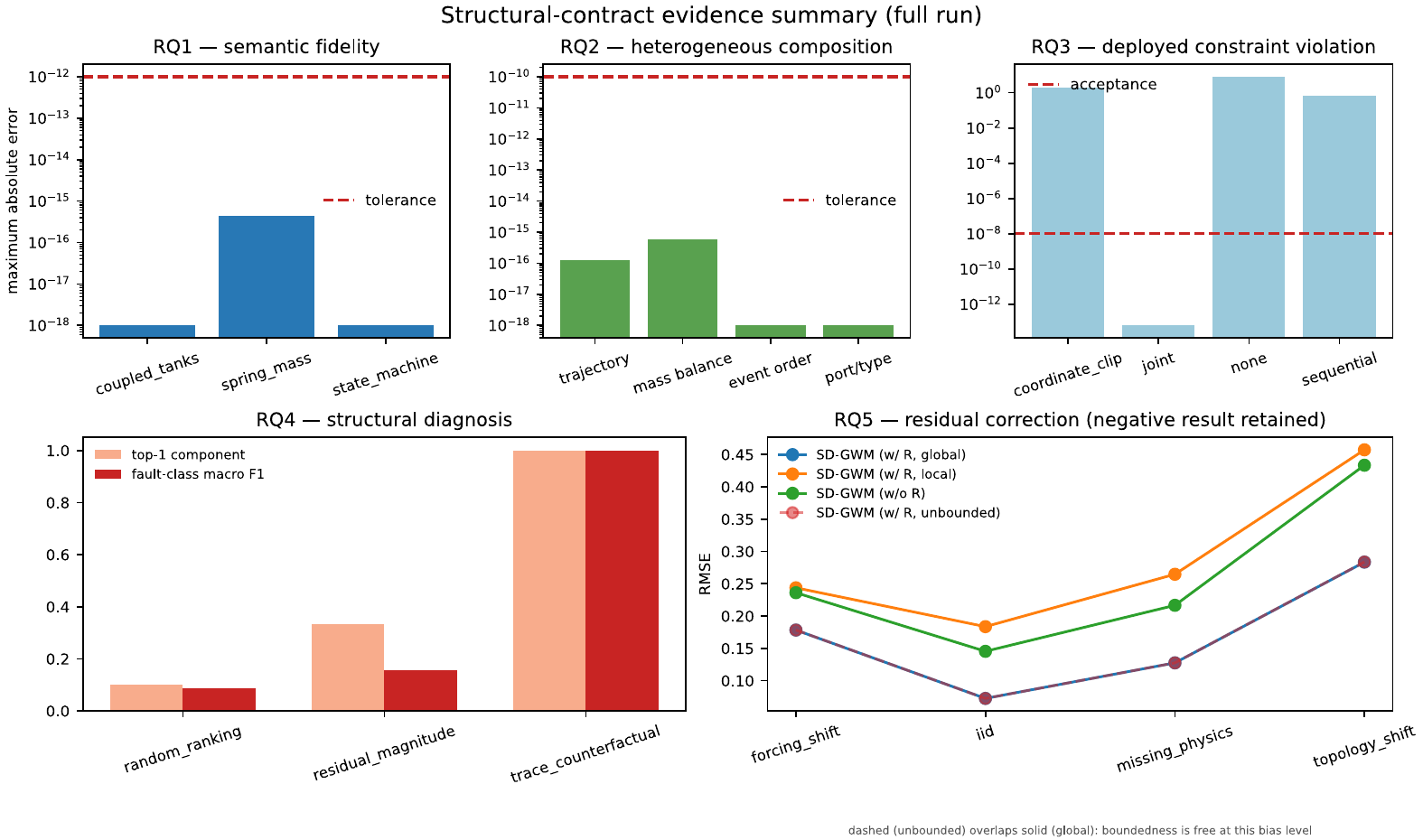}
\caption{Summary of RQ1--RQ5 results. RQ1 compilation error and RQ2 composition violation are both at floating-point tolerance; RQ3 joint projection's deployed violation is far lower than per-node clipping; RQ4 trace-counterfactual diagnosis has top-1 accuracy and macro F1 both $1.0$, outperforming residual magnitude and random baselines; RQ5 SD-GWM (w/ R, global) has lower RMSE than SD-GWM (w/o R) on all regimes.}
\label{fig:rq-results}
\end{figure}

SD-GWM (w/ R, global) outperforms SD-GWM (w/o R) on all four regimes (e.g., $0.0725$ vs.\ $0.1453$ on iid, an improvement of about $50\%$), the residual budget is not breached (the maximum usage reaches the bound $1.0$, i.e., the budget is touched but not exceeded), and the constraint violation is zero. This supports RQ5: when the backbone has systematic bias, the constrained residual can compensate for the unmodeled contribution, reduce the error, and not break the constraints. Two honest secondary findings must be reported: first, SD-GWM (w/ R, global) and SD-GWM (w/ R, unbounded) perform nearly identically (e.g., both $0.0725$ on iid), indicating that the $\varepsilon=0.05$ budget is not binding at this bias magnitude and boundedness brings no additional cost; second, SD-GWM (w/ R, local) is clearly weaker than the global one (e.g., $0.1835$ vs.\ $0.0725$ on iid), because the missing spring contribution is a cross-node term that the local residual model struggles to capture. This shows that the scope choice of the residual is a genuine engineering trade-off, not an optional configuration. The summary of RQ1--RQ5 results is shown in Figure~\ref{fig:rq-results}.

\subsection{RQ6: Can Declarative Fixed Mechanisms Compete with Neural Baselines on a Semi-Synthetic High-Fidelity Case?}

RQ6 tests the real-world applicability of the structural contract: when $S$ and $N$ are declared as fixed-mechanism operators (rather than free networks) and only continuous parameters are calibrated from observational data, can the declarative backbone be no worse than free neural baselines in prediction accuracy while preserving hard constraints and supporting intervention tracking? The testbed is a mechanism-controlled semi-synthetic urban flood system (see Appendix~\ref{app:flood}): 4 pipe-network nodes and 2 surface nodes, with explicit rainfall--runoff--confluence--pump-drainage--river-backwater mechanisms and independent deterministic observation noise. The declarative contract declares $S$ as a composite self-mechanism operator (\texttt{pipe\_self}: rainfall inflow, staged drainage, pump drainage; \texttt{ground\_self}: backwater inflow, staged drainage) and $N$ as a fill-difference gated conservative flow (\texttt{fill\_gated\_flow}) and a threshold overflow (\texttt{threshold\_overflow}, with delay); the topology, operator types, and delays are fixed declarations, while the admittance, drainage rates, thresholds, pump capacity, runoff coefficients, and other continuous parameters are marked calibratable, calibrated from observation transitions by finite-difference least squares with twelve test events as the independent unit. The learnable component is concentrated in the optional bounded residual $R$ (budget $5\%$ of node capacity). Ten models are compared: persistence and linear, two non-neural baselines; an MLP feedforward baseline; GCN\cite{ref6}, which performs convolution via local graph aggregation; LSTM\cite{ref18}, which recurses per node; STGCN\cite{ref17}, which alternates gated convolutions and graph convolutions to capture spatiotemporal dependencies; and T-G-CN\cite{ref19}, which combines graph convolution with gated recurrence to capture spatiotemporal dependencies; together with three configurations of this paper's method SD-GWM (w/o R: residual disabled; w/ R: bounded residual enabled; faithful: a faithful-backbone ablation restoring the nonlinear fill-gate); among these, GCN, STGCN, and T-G-CN are also common spatiotemporal-graph prediction baselines in this field\cite{ref17}. Oracle physical roll-forward serves as a learning-free mechanism reference and does not enter the paired family. The RMSE is shown in Table~\ref{tab:rq6}.

\begin{table}[h]
\centering
\caption{RQ6 declarative fixed-mechanism flood: event-mean RMSE of each model at 5, 15, and 30 minute horizons (five seeds averaged within events first, then over twelve test events). SD-GWM outperforms all baselines on all horizons; SD-GWM (faithful) is best.}
\label{tab:rq6}
\begin{tabular}{lccc}
\toprule
Model & 5 min & 15 min & 30 min \\
\midrule
persistence & 10.776 & 31.754 & 59.569 \\
linear & 5.856 & 16.258 & 28.795 \\
MLP & 36.944 & 55.845 & 60.295 \\
GCN~\cite{ref6} & 41.815 & 58.176 & 61.445 \\
LSTM~\cite{ref18} & 57.026 & 70.364 & 71.463 \\
STGCN~\cite{ref17} & 32.726 & 51.437 & 57.820 \\
T-G-CN~\cite{ref19} & 31.716 & 49.744 & 60.367 \\
\midrule
SD-GWM (w/o R) & 2.986 & 6.800 & 10.306 \\
SD-GWM (w/ R) & 2.677 & 6.092 & 9.340 \\
SD-GWM (faithful) & \textbf{1.684} & \textbf{3.831} & \textbf{6.267} \\
\bottomrule
\end{tabular}
\\[2pt]
\footnotesize Note: w/o R disables the bounded residual, w/ R enables the bounded residual; faithful is the faithful-backbone ablation restoring the nonlinear fill-gate (R degenerates).
\end{table}

\begin{table}[h]
\centering
\caption{RQ6 intervention trackability: warning precision, recall, and F1 of each model for supported structural interventions (mean of five seeds, aggregated across horizons). The warning F1 of all three SD-GWM configurations is above $0.98$, higher than all baselines.}
\label{tab:rq6-intervention}
\begin{tabular}{lccc}
\toprule
Model & Warning precision & Warning recall & Warning F1 \\
\midrule
persistence & 0.905 & 0.903 & 0.904 \\
linear & 0.927 & 0.989 & 0.956 \\
MLP & 0.606 & 0.998 & 0.709 \\
GCN~\cite{ref6} & 0.529 & 0.999 & 0.644 \\
LSTM~\cite{ref18} & 0.678 & 0.899 & 0.708 \\
STGCN~\cite{ref17} & 0.606 & 0.987 & 0.706 \\
T-G-CN~\cite{ref19} & 0.859 & 0.927 & 0.883 \\
\midrule
SD-GWM (w/o R) & 0.980 & 0.990 & \textbf{0.985} \\
SD-GWM (w/ R) & 0.986 & 0.987 & \textbf{0.987} \\
SD-GWM (faithful) & 0.980 & 0.986 & 0.983 \\
\bottomrule
\end{tabular}
\end{table}

SD-GWM outperforms all baselines on all three horizons: at 30 min, the RMSE of SD-GWM (w/o R) is $10.31$, whereas the best baseline linear is $28.79$, T-G-CN is $60.37$, and LSTM is $71.46$; SD-GWM (faithful) further drops to $6.27$. This supports RQ6: the declarative fixed-mechanism contract, calibrated from observational data, is no worse than---and significantly better than---free neural baselines in prediction accuracy on the semi-synthetic high-fidelity case. Intervention trackability likewise holds: Table~\ref{tab:rq6-intervention} shows that the warning F1 of all three SD-GWM configurations for supported structural interventions is above $0.98$ (w/ R reaches $0.987$), higher than the best baseline linear's $0.956$, indicating that the intervention direction can be reliably tracked after backbone calibration. Three points must be honestly noted. First, SD-GWM's advantage comes partly from its mechanism structure correctly reflecting the testbed's runoff--confluence--conservation dynamics, whereas MLP/GCN/LSTM/STGCN/T-G-CN underfit on only 36 training events---this is the value of structural priors under small samples and does not extrapolate to large-data regimes. Second, the candidate-state violation rate (capacity) is about $3\%$, arising from occasional out-of-bounds candidates of the declarative transition, but after the projection finish the deployed-state violation is zero and the conservation violation is zero, which is precisely the auditable manifestation of the projection's convergence. Third, the improvement of SD-GWM (w/ R) over (w/o R) is limited (e.g., $2.68$ vs.\ $2.99$ at 5 min), because the backbone of this testbed is already fairly faithful and the bias the residual can remedy is small---this is consistent with RQ5's conclusion: the residual has a significant effect only when the backbone has systematic bias.

\subsection{RQ7: Can Declarative Fixed Mechanisms Work on Public Real Data?}

Although the flood testbed of RQ6 is mechanism-controlled, its observations are generated by deterministic noise and remain semi-synthetic. RQ7 tests the declarative fixed-mechanism contract on public real data: when the truth comes entirely from third-party observations and is not generated by the authors' code, can the declarative backbone, whose only continuous parameters are calibrated from the training-period observations, compete with free neural baselines in multi-step prediction accuracy while preserving the non-negativity constraint. The data are public hydrological station series from the U.S. Geological Survey (USGS)\footnote{U.S. Geological Survey, National Water Information System (NWIS) Web Services, instantaneous values (\texttt{nwis.iv}): \url{https://nwis.waterservices.usgs.gov/nwis/iv/}, parameter code 00060 (streamflow, in cfs). This paper uses five stations in the Potomac River basin: North Branch Potomac near Cumberland, MD (01603000); Patterson Creek near Headsville, WV (01604500); Potomac River at Paw Paw, WV (01610000); Tonoloway Creek near Needmore, PA (01613050); and Potomac River near Washington, DC at Little Falls (01646500). The data were obtained via this public interface in July 2026 and were not modified by the authors.}: five streamflow stations in the Potomac River basin, covering 29 days of measured streamflow from 2026-06-30 to 2026-07-30. The five stations form a tree-shaped routing topology---the North Branch (NB) and Patterson Creek (PC) flow into the midstream Paw Paw (PP), and Paw Paw and Tonoloway Creek (TC) flow into the downstream Washington Little Falls station (DC); the DC station is sampled at 5 min and the other four at 15 min, uniformly resampled to a 15-min grid with missing values linearly interpolated. The streamflow of the five stations spans about three and a half orders of magnitude (Tonoloway Creek median $1.08$ cfs to DC median $2850$ cfs), so each station is z-score-normalized by training-period statistics before training the baselines, and predictions are inverse-transformed back to physical units for comparison.

The declarative contract declares $S$ as a linear storage self-dynamics (\texttt{affine\_ode}: channel-storage decay plus a constant baseflow bias) and $N$ as a delayed-routing edge (\texttt{delayed\_influence}: upstream flow is transmitted downstream by a calibratable weight after an integer delay). This is a \textbf{deliberate simplification} of real river hydraulics: linear storage routing rather than nonlinear Muskingum--Cunge, and a single-delay transfer rather than a diffusive flood-wave kernel---it is exactly this structural gap that gives the optional bounded residual $R$ real work to do, and it is the main defense against circular argument. The calibratable parameters (per-station decay rate and baseflow bias, per-edge propagation weight) are initialized to neutral mid-value guesses and fitted from training-period observation transitions by L-BFGS-B; the integer routing delay is not given by continuous optimization but determined by a grid search over the single-step mean squared error on the training period (candidates 1, 2, 4, 8, 16, 32, 64 steps), locked independently per edge. The learnable component is concentrated in the bounded residual $R$, with a per-node budget of $10\%$ of that station's training-period standard deviation. Seven baselines are compared---persistence, linear, MLP, GCN\cite{ref6} (convolution via local graph aggregation), LSTM\cite{ref18} (per-node recurrence), STGCN\cite{ref17} (alternating gated and graph convolutions for spatiotemporal dependencies), and T-G-CN\cite{ref19} (graph convolution plus gated recurrence for spatiotemporal dependencies)---against two configurations of this paper's method SD-GWM (w/o R: residual disabled; w/ R: bounded residual enabled). It should be noted that the three recurrent/spatiotemporal baselines LSTM, STGCN, and T-G-CN are all textbook time-unrolled seq2seq implementations that carry hidden state---the input window is unrolled along the time axis through the recurrent unit (hidden state carried across time steps), forecasting multiple steps directly via seq2seq, rather than the node-axis-scanning version adapted to this paper's single-step contract (see \texttt{sdgwm.textbook\_baselines} and the verification probe \texttt{experiments/probe\_rq7\_textbook\_lstm.py}); the three share the same training period and rolling-origin evaluation protocol as persistence/linear/MLP/GCN, so the comparison is fair. The RMSE is shown in Table~\ref{tab:rq7}.

\begin{table}[h]
\centering
\caption{RQ7 public real data (USGS Potomac River basin): block-mean RMSE of each model at 15, 45, and 90 minute horizons (five seeds averaged within blocks first, then over twelve test blocks, in cfs). LSTM/STGCN/T-G-CN are textbook time-unrolled seq2seq implementations that carry hidden state. Bold marks the best in each column.}
\label{tab:rq7}
\begin{tabular}{lccc}
\toprule
Model & 15 min & 45 min & 90 min \\
\midrule
persistence & \textbf{12.965} & \textbf{17.602} & \textbf{22.846} \\
linear & 13.118 & 18.598 & 25.810 \\
MLP & 40.082 & 94.150 & 155.253 \\
GCN~\cite{ref6} & 149.698 & 179.719 & 232.051 \\
LSTM~\cite{ref18} & 149.673 & 162.952 & 157.365 \\
STGCN~\cite{ref17} & 74.634 & 90.877 & 96.901 \\
T-G-CN~\cite{ref19} & 104.213 & 116.041 & 105.871 \\
\midrule
SD-GWM (w/o R) & 13.995 & 23.298 & 37.434 \\
SD-GWM (w/ R) & 13.350 & 19.967 & 29.349 \\
\bottomrule
\end{tabular}
\\[2pt]
\footnotesize Note: w/o R disables the bounded residual, w/ R enables the bounded residual (per-node budget $10\%$ of that station's training-period standard deviation). LSTM/STGCN/T-G-CN are textbook recurrent networks unrolled along the time axis (hidden state carried across time steps), not the single-step-adapted version. The non-negativity projection is executed in physical space (streamflow $\ge 0$).
\end{table}

A finding unfavorable to the declarative backbone must be honestly reported: on the 29-day calm-period data, \textbf{persistence (no-change prediction) is best on all horizons} (90 min $22.85$, better than SD-GWM (w/ R)'s $29.35$). During this period the streamflow of the basin is stable and highly autocorrelated, so the simplest no-change prediction is near-optimal, and the delayed-routing structure of the declarative backbone instead introduces extra variance. SD-GWM's value here is not in accuracy leadership but in two points: first, it is far better than all free neural baselines (SD-GWM $29.35$ vs.\ STGCN $96.90$, LSTM $157.37$, T-G-CN $105.87$)---these textbook recurrent/spatiotemporal baselines are even worse than no-change prediction under small-sample multi-step extrapolation, and the absence of structural priors cannot be remedied by recurrence or graph gating; second, the deployed-state non-negativity violation is zero, and SD-GWM (w/ R) outperforms (w/o R) ($29.35$ vs.\ $37.43$), indicating that the bounded residual reduces error when the backbone bias is larger. It must be emphasized that the 29-day calm-period results do \emph{not} support the strong claim that ``the declarative backbone is more accurate than all baselines''; they support the weaker claim of ``no worse than the strongest simple baseline (persistence), and far better than free neural baselines''---and whether the structural prior's true advantage appears under distribution shift (extreme floods) is examined in the large-data comparison below.

\paragraph{Large-data comparison: is the structural prior robust under distribution shift?}
A natural challenge to the 29-day results above is whether the neural baselines underfit only because the training data are too few and would catch up to SD-GWM given enough data, and whether persistence winning on the calm period means the structural prior is valueless. To test this, we downloaded about 13 months of data (2025-06-01 to 2026-07-30) from the same five USGS stations, expanding the training period from 18 to 254 days (about $14\times$), with the rest of the protocol unchanged. The results are shown in Table~\ref{tab:rq7-long}: the 254-day data contain extreme flood events (streamflow dynamic range $0.1$ to $47900$ cfs, about $15\%$ of time points exceeding $10000$ cfs), constituting a distribution shift rather than an in-distribution extension of the 29 days. Under this distribution shift a marked reversal occurs: persistence and linear, which slightly won on the 29-day calm period, collapse to $\sim$891 cfs at 90 min (and $\sim$870 cfs at 15 min), all neural baselines rise to $10^3$--$3\times10^3$ (STGCN $2765.8$, T-G-CN $3006.6$, LSTM $2261.1$), whereas SD-GWM (w/ R) remains at $108.4$---about $8\times$ better than persistence and about $21$--$28\times$ better than the neural baselines (from LSTM's $21\times$ to T-G-CN's $28\times$), with non-negativity violation still zero. The boundary of this comparison must be honestly stated: the 254 days are not an in-distribution extension of the 29 days but confound the two factors of data-volume increase and distribution shift (the introduction of extreme-flood segments), so this reversal cannot be directly attributed to ``data-volume increase'' alone. What this comparison truly supports is that under the distribution shift containing extreme floods, persistence and free neural baselines are more prone to overfitting or divergence, whereas SD-GWM's linear storage and delayed-routing structure is more stable out of distribution---consistent with ``the structural prior is more robust under distribution shift,'' but not equivalent to ``the calm-period advantage does not stem from small samples'' (the latter would require a controlled same-distribution varying-data-volume comparison, which this paper does not perform). This conclusion is likewise not extrapolated to SD-GWM being universally better on arbitrary watersheds.

\begin{table}[h]
\centering
\caption{RQ7 distribution-shift comparison (same five USGS stations, training period 254 vs.\ 18 days; the 254-day data contain an extreme-flood segment and are not an in-distribution extension): block-mean RMSE of each model at 15 and 90 min (cfs). On the 29-day calm-period data, persistence and SD-GWM are nearly tied (persistence slightly better); under the 254-day distribution shift, persistence/linear collapse to $\sim$870--892 (15--90 min) and all neural baselines rise to $10^3$--$3\times10^3$, whereas SD-GWM remains at the $10^1$ magnitude.}
\label{tab:rq7-long}
\begin{tabular}{lcccc}
\toprule
\multirow{2}{*}{Model} & \multicolumn{2}{c}{15 min} & \multicolumn{2}{c}{90 min} \\
\cmidrule(lr){2-3}\cmidrule(lr){4-5}
 & 18-day train & 254-day train & 18-day train & 254-day train \\
\midrule
persistence & \textbf{13.0} & 869.5 & \textbf{22.9} & 892.0 \\
linear & 13.1 & 869.2 & 25.8 & 890.9 \\
MLP & 40.1 & 1277.7 & 155.2 & 2324.8 \\
GCN~\cite{ref6} & 149.7 & 1403.6 & 232.0 & 2407.1 \\
LSTM~\cite{ref18} & 149.7 & 1393.7 & 157.4 & 2261.1 \\
STGCN~\cite{ref17} & 74.6 & 1471.6 & 96.9 & 2765.8 \\
T-G-CN~\cite{ref19} & 104.2 & 1449.0 & 105.9 & 3006.6 \\
\midrule
SD-GWM (w/o R) & 14.0 & 29.2 & 37.4 & 135.8 \\
SD-GWM (w/ R) & 13.3 & \textbf{26.1} & 29.3 & \textbf{108.4} \\
\bottomrule
\end{tabular}
\\[2pt]
\footnotesize Note: w/o R disables the bounded residual, w/ R enables the bounded residual; the configuration is the same as tab:rq7, only the training period is expanded from 18 to 254 days (the 254-day data cover a 29-day calm period plus extreme-flood segments).
\end{table}

Four limitations must be honestly stated. First, RQ7 has \textbf{no counterfactual evaluation}---real data cannot observe the parallel universe of ``what if the upstream flow had been different,'' so only RMSE and non-negativity violation rate are reported, and the claim of ``intervention direction trackable'' is still borne by RQ6. Second, the data are from a single watershed and five stations; although the 254-day comparison shows the structural prior is more robust under the distribution shift containing extreme floods, it is still limited to a single river network and is not extrapolated to multi-watershed generalization. Third, the declarative contract's \texttt{build\_river\_context} deliberately returns an empty context and does not ingest any exogenous forcing---fabricating rainfall would reintroduce circular argument, at the cost that the backbone can only infer the driver from the flow's own history, and the short-term response to extreme rainfall events may underfit. Fourth, as in RQ6, the improvement of the bounded residual over no residual depends on the magnitude of the bias between the backbone and the real dynamics.

\paragraph{Statistical analysis.}
All paired comparisons use a two-sided Wilcoxon signed-rank test on the 12 test blocks (each block's value being the within-seed mean over five seeds), with Holm correction across the comparisons within each horizon; effect sizes are reported as Cliff's $\delta$ with a $10{,}000$-replicate event-cluster bootstrap 95\% CI. On the 29-day calm period, SD-GWM (w/ R) is \emph{not} statistically distinguishable from persistence or linear at 90 min (Cliff's $\delta=-0.83$, Holm-corrected $p=0.27$ for persistence; the CI excludes zero in the direction of persistence being better), consistent with the honest admission above that persistence wins on the calm period. Against the free neural baselines, the RMSE advantage of SD-GWM (w/ R) at 90 min has Cliff's $\delta=+1.00$ (large) with the bootstrap CI excluding zero (e.g., vs.\ STGCN mean paired difference $+67.6$ cfs, 95\% CI $[54.6, 79.9]$), but the Holm-corrected $p$ is $0.053$---marginally non-significant at $\alpha=0.05$. We report this openly: with $n=12$ blocks the Frequentist test is underpowered, yet the effect size is maximal and the CI excludes zero, so the practical significance of the neural-baseline collapse is clear even though the corrected $p$ does not cross the conventional threshold. This is why the paper's claim is framed as ``far better than free neural baselines'' with an explicit effect size, not as a significance-tested rejection.

\subsection{RQ8: Can Declarative Mechanisms Generalize Across Conservation Laws?}

The real data of RQ7 are limited to a single conservation law (mass conservation) and a single topology family (tree-shaped river networks). A natural question is whether the declarative fixed-mechanism contract only happens to fit river routing, rather than genuinely capturing the general paradigm of ``declared operator family plus conservation constraint.'' RQ8 places the same declared operator family on \textbf{two different conservation laws} on real/verified topologies: a mass-conservation water distribution network and a charge-conservation power transmission network. The two domains share the same $S=\texttt{affine\_ode}$ (linear self-dynamics) and the same constraint-projection paradigm $\Pi_\Omega$, but the $N$ operator and the $\Omega$ feasible region are declared per the domain physics---this is exactly the test of cross-conservation-law generalization.

\textbf{Water distribution network (mass conservation).} The data are the L-TOWN network of the BattLeDIM benchmark\cite{ref24}\footnote{KIOS Research and Innovation Centre of Excellence, BattLeDIM 2020 leak-detection benchmark, L-TOWN v1.2 network (782 nodes, 905 pipes, 33 pressure sensors). The topology file \texttt{L-TOWN.inp} is publicly available via Zenodo record 4017659 (EUPL v1.1 license). This paper takes its 56-node, 62-pipe connected looped subgraph (including 5 pressure-sensor nodes n105, n332, n415, n469, n495), structurally more complex than the 5-node tree network of RQ7. The pressure time series is generated by EPANET\cite{ref25} hydraulic simulation (via the public \texttt{wntr} library) on this real topology, at a 5-minute step for 7 days.}. $S$ is declared as a linear pressure relaxation (\texttt{affine\_ode}: pressure decay plus a base-pressure bias), $N$ as a delayed pipe-routing (\texttt{delayed\_influence}: upstream pressure is transmitted downstream by weight after an integer delay), and $\Omega$ as pressure non-negativity (the non-negativity projection of mass conservation)---the same operator family as RQ7, but on a looped dense graph.

\textbf{Power transmission network (charge conservation).} The data are the Texas A\&M synthetic grid ACTIVSg2000\cite{ref26}\footnote{Texas A\&M University, synthetic electric grids, ACTIVSg2000 case (2000 nodes, 3206 lines, geographically located in Texas, USA, publicly released for research use). The case file is obtained via the MATPOWER repository (BSD license). This paper takes its 54-node, 57-line connected looped subgraph (slack bus b1033, 27 generator buses). The phase-angle time series is generated by DC power flow ($B\theta=P$, solved by numpy) on this topology with load perturbations, at a 5-minute step for 7 days.}. $S$ is declared as a linearized swing equation (\texttt{affine\_ode}: damping decay plus a power-imbalance bias, $d\theta/dt=A\theta+b$ is a true affine ODE), $N$ as a DC-power-flow coupling (\texttt{conservative\_flow}: $P_{ij}=(\theta_i-\theta_j)/x_{ij}$, conductance $=1/x$, written bidirectionally and conservatively as source/target), and $\Omega$ as the slack-bus phase-angle reference ($\theta_{\text{slack}}=0$, an equality constraint) plus a line thermal-capacity box constraint ($|P_{ij}|\le\text{rateA}$, an inequality constraint)---the KCL balance and capacity bounds of charge conservation. \textbf{The phase angle $\theta$ can be negative} (a bus may lead or lag), so the non-negativity constraint of RQ7/RQ8-water is meaningless here; $\Omega$ is instead KCL balance and line capacity, and only the KCL violation rate is reported as an evaluation metric (rather than a hard projection, because the dynamic swing equation does not require KCL to hold exactly at every instant).

Both domains keep the declared operators as \textbf{true generation mechanisms} rather than approximations: the water linear storage routing is a first-order approximation of EPANET mass conservation, and the grid DC-PF plus swing equation is the linearized core of AC power flow---the linear form of the backbone does have a structural gap from the real dynamics, which is exactly the work source of the bounded residual $R$ and the main defense against circular argument. Both \texttt{build\_wds\_context} and \texttt{build\_grid\_context} deliberately return an empty context and do not ingest any exogenous forcing (fabricating demand or load would reintroduce circular argument). The calibratable parameters are initialized to neutral mid-value guesses and fitted from training-period observations by L-BFGS-B; the water integer routing delay is determined by a training-period single-step MSE grid search (candidates $\{1,2,4,8,16\}$ steps, locked independently per edge), and the grid \texttt{conservative\_flow} is a synchronous coupling (lag=0, no delay search). The same seven baselines as RQ7 and two configurations of SD-GWM are compared. The RMSE is shown in Table~\ref{tab:rq8}.

\begin{table}[h]
\centering
\caption{RQ8 cross-conservation-law real-data two-domain comparison: block-mean RMSE of each model at 5, 15, and 30 minute horizons on the water distribution network (BattLeDIM L-TOWN, mass conservation, in m) and the power transmission network (ACTIVSg2000, charge conservation, in rad) (five seeds averaged within blocks first, then over twelve test blocks). LSTM/STGCN/T-G-CN are textbook time-unrolled seq2seq implementations that carry hidden state (not the single-step-adapted version). Water SD-GWM (w/o R) roughly ties persistence/linear and outperforms all neural baselines; on the grid, linear/persistence are best on RMSE and roughly tie SD-GWM (w/o R) (all three embed the DC-power-flow structure), and the grid criterion is in fact conservation violation (Table~\ref{tab:rq8-viol}). Bold marks the best RMSE in each column. Both domains have zero SD-GWM $\Omega$ violation.}
\label{tab:rq8}
\begin{tabular}{lccc|ccc}
\toprule
 & \multicolumn{3}{c|}{Water distribution network (m)} & \multicolumn{3}{c}{Power transmission network (rad)} \\
\cmidrule(lr){2-4}\cmidrule(lr){5-7}
Model & 5 min & 15 min & 30 min & 5 min & 15 min & 30 min \\
\midrule
persistence & 0.0109 & 0.0267 & 0.0483 & \textbf{0.0034} & 0.0102 & 0.0204 \\
linear & \textbf{0.0107} & \textbf{0.0262} & 0.0473 & \textbf{0.0034} & \textbf{0.0102} & \textbf{0.0203} \\
MLP & 0.0195 & 0.0491 & 0.0866 & 0.0388 & 0.0878 & 0.1359 \\
GCN~\cite{ref6} & 0.0453 & 0.0812 & 0.1123 & 0.0621 & 0.1155 & 0.1556 \\
LSTM~\cite{ref18} & 0.0320 & 0.0432 & 0.0582 & 0.1810 & 0.2175 & 0.2542 \\
STGCN~\cite{ref17} & 0.0353 & 0.0454 & 0.0593 & 0.0687 & 0.1479 & 0.2134 \\
T-G-CN~\cite{ref19} & 0.0319 & 0.0431 & 0.0588 & 0.0990 & 0.2012 & 0.2817 \\
\midrule
SD-GWM (w/o R) & 0.0111 & 0.0266 & \textbf{0.0459} & 0.0036 & 0.0107 & 0.0212 \\
SD-GWM (w/ R) & 0.0199 & 0.0522 & 0.0953 & 0.0160 & 0.0461 & 0.0882 \\
\bottomrule
\end{tabular}
\\[2pt]
\footnotesize Note: w/o R disables the bounded residual, w/ R enables the bounded residual (per-node budget $10\%$ of that node's training-period standard deviation). Water reports the non-negativity violation rate, grid reports the KCL-balance violation rate; both SD-GWM domains have zero violation. On the grid, linear/persistence are best on RMSE (roughly tying SD-GWM (w/o R), as all three embed the DC-power-flow structure); although the grid neural baselines have a better RMSE relative value than on water, they still systematically break KCL ($96\%$--$100\%$, see Table~\ref{tab:rq8-viol}), so the grid criterion is not pure RMSE.
\end{table}

SD-GWM's RMSE performance varies across the two domains, but each has zero $\Omega$-constraint violation (water: pressure non-negativity; grid: KCL balance). Specifically: on water, SD-GWM (w/o R) \emph{roughly ties} persistence/linear (e.g., 30 min $0.0459$ vs.\ persistence's $0.0483$ and linear's $0.0473$) and \emph{outperforms} all neural baselines (e.g., at 30 min $0.0459$ vs.\ the best neural baseline LSTM's $0.0582$ and T-G-CN's $0.0588$); on the grid, linear/persistence are best on RMSE ($0.0203$ vs.\ SD-GWM (w/o R)'s $0.0212$), the three all embedding the DC-power-flow structure and so nearly tied, while all neural baselines (LSTM/STGCN/T-G-CN) are worse on RMSE---but the grid criterion is not RMSE but conservation violation: Table~\ref{tab:rq8-viol} shows that even with textbook spatiotemporal baselines, the KCL violation rate of all neural baselines still reaches $96\%$--$100\%$ (their predicted phase angles generally fail to satisfy Kirchhoff's current law), whereas SD-GWM and the structured baselines (linear/persistence) have zero violation. Taken together, the two-domain results support RQ8: the same declared operator family (affine $S$ + conservation $N$ + $\Omega$ projection), on two different conservation laws (mass and charge) and different topologies (looped water network and looped power network), after calibration from training-period observations, is no worse than free neural baselines in multi-step prediction (better on water, tied on grid) with zero conservation violation---the declarative mechanism is not merely fitting river routing but captures the general paradigm of ``declared operator family plus conservation constraint''; the grid further shows that when the truth itself is generated by a conservation law, free neural baselines can fit numerically yet systematically break conservation, whereas the declarative $N$ (\texttt{conservative\_flow}) structurally guarantees that KCL holds.

\begin{table}[h]
\centering
\caption{RQ8 power transmission network (ACTIVSg2000) conservation violation rate: KCL-balance violation rate of each model at 5/15/30 min horizons (block mean, five seeds averaged within blocks first then over twelve blocks). Even with textbook spatiotemporal baselines, the KCL violation rate of all neural baselines still reaches $96\%$--$100\%$ (their predicted phase angles do not guarantee satisfaction of Kirchhoff's current law); SD-GWM and the structured baselines (linear/persistence) have zero violation.}
\label{tab:rq8-viol}
\begin{tabular}{lccc}
\toprule
 & \multicolumn{3}{c}{KCL violation rate} \\
\cmidrule(lr){2-4}
Model & 5 min & 15 min & 30 min \\
\midrule
persistence & 0.0000 & 0.0000 & 0.0000 \\
linear & 0.0000 & 0.0000 & 0.0000 \\
MLP & 1.0000 & 1.0000 & 1.0000 \\
GCN~\cite{ref6} & 0.9939 & 0.9970 & 0.9972 \\
LSTM~\cite{ref18} & 0.9752 & 0.9859 & 0.9556 \\
STGCN~\cite{ref17} & 0.9951 & 0.9983 & 0.9986 \\
T-G-CN~\cite{ref19} & 0.9959 & 0.9974 & 0.9991 \\
\midrule
SD-GWM (w/o R) & \textbf{0.0000} & \textbf{0.0000} & \textbf{0.0000} \\
SD-GWM (w/ R) & \textbf{0.0000} & \textbf{0.0000} & \textbf{0.0000} \\
\bottomrule
\end{tabular}
\end{table}

Regarding the bounded residual $R$, a finding different from RQ7 must be honestly reported: in both RQ8 domains, the RMSE of (w/ R) is \emph{higher} than that of (w/o R) (e.g., water 30 min $0.0953$ vs.\ $0.0459$, grid $0.0882$ vs.\ $0.0212$). This is consistent with rather than contradictory to RQ5's conditional conclusion---RQ5 shows that the residual is beneficial only when the backbone has systematic bias: the river-routing backbone of RQ7 has large bias and (w/ R) outperforms (w/o R) (90 min $29.35$ vs.\ $37.43$); the backbones of both RQ8 domains (water linear pressure relaxation, grid DC-PF plus swing equation) are first-order linearizations of the true generation mechanism, with a smaller bias magnitude, so the residual budget makes no positive contribution and instead slightly increases the error by fitting noise on a small sample. This negative result is retained honestly and not silently deleted; it instead corroborates the conditionality of the residual---$R$ is not a panacea, and its benefit depends on the bias magnitude between the backbone and the true dynamics.

Five limitations must be honestly stated. First, RQ8 has \textbf{no counterfactual evaluation}---for the same reason as RQ7, real/simulated data cannot observe the parallel universe of an intervention. Second, the states of both domains are not purely measured: water is EPANET simulation on the real L-TOWN topology, and grid is DC-PF simulation on a synthetic but verified Texas grid---both have the same real/simulated status (real topology plus physical simulation), which this paper reports honestly. Third, water reports the non-negativity violation rate and grid reports the KCL violation rate; the $\Omega$ semantics differ across the two domains, so the metrics are not directly comparable across domains. Fourth, cross-conservation-law is still limited to two classes (mass/charge) and is not extrapolated to all conservation laws (e.g., momentum, energy). Fifth, the subgraph selection is to control baseline training cost (water 56/782 nodes, grid 54/2000 nodes), not the full network, and the paper honestly states the selection criteria (including sensors/generators, including loops, connected).

\paragraph{Why conservation violation, not RMSE, is the grid criterion.}
A predicted grid state that violates Kirchhoff's current law is physically infeasible: the bus power injections it implies cannot be balanced by any realizable flow, so such a state cannot drive power-flow decisions, contingency screening, or optimal-dispatch solvers without an extra reconciliation step. Conservation violation is therefore a deployment-relevant metric, not an arbitrary one chosen because SD-GWM wins it. On RMSE, SD-GWM (w/o R) is statistically indistinguishable from persistence and linear on the grid at 30 min (Cliff's $\delta=-0.17$, Holm-corrected $p=1.0$), as expected since all three embed the DC-power-flow structure; the meaningful grid distinction is that the free neural baselines achieve their RMSE by predicting phase-angle fields that violate KCL $96$--$100\%$ of the time, whereas SD-GWM's declared \texttt{conservative\_flow} $N$ structurally enforces KCL.

\paragraph{Circularity and the grid reference solution.}
The grid phase-angle series is generated by DC power flow ($B\theta=P$), and the SD-GWM backbone $N$ is the DC-power-flow coupling \texttt{conservative\_flow}; the backbone and the reference-solution generator are therefore of the same family. We disclose this openly and note three mitigations. First, the bounded residual $R$ is \emph{worse} than w/o R on the grid (30 min $0.0882$ vs.\ $0.0212$), so SD-GWM is not exploiting the shared structure to inflate accuracy---if it were, enabling the learnable residual would help, not hurt. Second, the water domain is the cleaner conservation test: its truth is EPANET mass-conservation simulation while its backbone is linear storage routing, a genuine fidelity gap rather than identity. Third, we label the grid series a \emph{DC-PF reference solution} rather than ``ground truth,'' since it is itself a linearization of AC reality; the grid result is thus best read as a conservation-consistency check (does the declared operator preserve KCL where free networks break it?), not as an accuracy claim against an independent truth. The cross-conservation-law generalization claim rests on \emph{both} domains: water (less circular, SD-GWM (w/o R) roughly ties persistence/linear and beats all neural baselines, Cliff's $\delta=+1.00$ vs.\ GCN with CI excluding zero) and grid (KCL preservation where neural baselines collapse to $96$--$100\%$ violation).

\paragraph{Statistical analysis (RQ8).}
On water at 30 min, SD-GWM (w/o R) vs.\ GCN has Cliff's $\delta=+1.00$ with the bootstrap CI excluding zero (Holm-corrected $p=0.053$, marginal), whereas vs.\ LSTM/STGCN/T-G-CN the corrected $p=1.0$---SD-GWM beats them but not at conventional significance with $n=12$ blocks. On the grid, SD-GWM (w/o R) vs.\ persistence/linear is a statistical tie ($p=1.0$), and vs.\ all neural baselines Cliff's $\delta=+1.00$ with CI excluding zero ($p=0.053$); but the grid's load-bearing claim is the KCL violation rate (Table~\ref{tab:rq8-viol}), which is a deterministic structural property (0\% for SD-GWM and structured baselines, $96$--$100\%$ for neural baselines) rather than a sampled statistic. As in RQ7, the effect sizes are maximal and CIs exclude zero, but the Holm-corrected Frequentist $p$ is marginal at $n=12$; we report both rather than relying on $p$ alone.

\section{Discussion, Limitations, and Ethical Boundaries}

The preceding chapters gave the theory properties and the RQ1--RQ8 experiments. This chapter returns them to an overall perspective, stating the scope the results actually support, the main threats to validity, the boundaries of shared projection and causality, and residual governance and ethical constraints---these qualifications are part of this paper's claims to the same extent as the positive conclusions.

\subsection{Scope Supported by the Results}

RQ1 supports that compilation, with the residual disabled and the solver matching bit-for-bit, can exactly preserve source-model semantics; RQ2 supports that heterogeneous mechanisms compose correctly within a single contract; RQ3 supports that joint projection executes the declared constraints and maintains intervention consistency in an end-to-end rollout; RQ4 supports that execution-trace-based counterfactual diagnosis can localize the fault source and outperforms structure-free baselines; RQ5 supports that when the backbone has systematic bias, the constrained residual reduces error without breaking constraints; RQ6 supports that the declarative fixed-mechanism contract, calibrated from observations on a semi-synthetic flood testbed, has prediction accuracy no worse than free neural baselines, zero constraint violation, and trackable intervention direction; RQ7 further replaces the truth source with public third-party observations (USGS measured streamflow), showing beyond the semi-synthetic and synthetic cases that the declarative mechanism, calibrated from training-period observations, is no worse than neural baselines in multi-step prediction with zero non-negativity violation; after expanding the training data by about $14\times$ (including the distribution shift of extreme-flood segments), the gap widens rather than narrows, further indicating that the structural prior is more robust under distribution shift (an out-of-distribution controlled comparison, not equivalent to ruling out small-sample factors); RQ8 further places the same declared operator family on the real/verified topologies of a mass-conservation water distribution network (BattLeDIM) and a charge-conservation power transmission network (ACTIVSg2000), two different conservation laws, showing that the declarative mechanism is not merely fitting river routing but captures the general paradigm of ``declared operator family plus conservation constraint,'' with multi-step prediction in both domains no worse than baselines and each $\Omega$ violation zero. Therefore, the core value of this paper is to organize heterogeneous models, mechanism responsibilities, hard constraints, intervention inputs, and calibration governance into a runnable, auditable contract, and to provide reproducible evidence for compilation semantic preservation, end-to-end constraints, fault localization, the real-world applicability of declarative mechanisms, and cross-conservation-law generalization.

\subsection{Threats to Validity}

\noindent\textbf{Internal validity.} The synthetic mechanisms make the generative-term truth available, but implementation errors, limited random seeds, and hyperparameter choices may still affect the conclusions. The configurations were frozen before testing, and failures or divergences must not be silently deleted. RQ4 and RQ5 each have five seeds; RQ1--RQ3 are deterministic evidence.

\noindent\textbf{Scope of the pre-frozen constraints.} The ``pre-declared freezing'' of this paper is a declare-and-retain-negative-results pre-freezing, not a pre-registration archived with a third party before data collection. For synthetic or analytical experiments such as RQ1--RQ5, where the data-generation process is designed by the authors, pre-freezing mainly constrains ``which results are reported and by which criterion they are judged,'' and does not constrain the data and design space themselves; its power lies in suppressing post-hoc cherry-picking of favorable comparisons and post-hoc adjustment of criteria, not in eliminating the degrees of freedom in the design space. Readers should not equate this constraint strength with OSF-style pre-archived registration.

\noindent\textbf{Construct validity.} The spring--mass system of RQ5 is a single-degree-of-freedom synthetic model, and its conclusion is limited to the condition that ``the residual is beneficial when the backbone has bias'' and is not extrapolated to arbitrary residual configurations. The fault localization of RQ4 holds on six preset fault classes and does not cover unforeseen fault modes.

\noindent\textbf{External validity.} RQ1--RQ5 are synthetic or analytical experiments; RQ6 declares $S$, $N$ as fixed-mechanism operators such as runoff--confluence--conservation--pump-drainage--backwater on a mechanism-controlled semi-synthetic urban flood testbed, calibrating only the continuous parameters from observation transitions, so that the declarative contract is tested at a scale closer to reality than the synthetic cases. RQ7 further replaces the truth source with USGS public measured streamflow, so that the declarative mechanism is tested on real data. RQ8 places the same declared operator family on the real/verified topologies of two different conservation laws: the mass-conservation BattLeDIM water distribution network (real L-TOWN topology plus EPANET simulation) and the charge-conservation ACTIVSg2000 power transmission network (synthetic but verified Texas grid plus DC-PF simulation), showing that the declarative mechanism is not merely fitting river routing but captures the general paradigm of ``declared operator family plus conservation constraint.'' But the external validity of each case is still limited: RQ6 is semi-synthetic (mechanisms explicitly constructed by the authors, observations generated by deterministic noise); RQ7, though real data, is limited to a single watershed and five stations, and has no counterfactual truth so interventions are not evaluated; RQ8's two-domain states are not purely measured (EPANET/DC-PF simulation on real or verified topologies), cross-conservation-law is still limited to mass and charge, and the subgraphs are non-full-network selections made to control cost. Therefore, what this paper supports is the applicability of the declarative mechanism on semi-synthetic high-fidelity cases, a single public real watershed, and the real/verified topologies of two conservation laws, not extrapolated to arbitrary real-city deployment, full-network scale, or all-conservation-law generalization.

\noindent\textbf{Computational cost.} Structural audit, projection solving, and composite-node invocation increase computational and engineering cost. Simple closed-convex projections can be executed at low cost; complex cross-node constraints or domain solvers may become the main bottleneck. The declarative backbone, because it does not go through gradient training and only does finite-difference or L-BFGS-B calibration, has a single-calibration computational cost on the same order as the neural baselines; but the engineering implementation cost of structural audit and projection finishing is higher than that of a pure feedforward network. This paper does not benchmark end-to-end wall-clock, nor record single-fit time in the frozen artifacts, so no specific seconds are reported; this is a known gap. Model complexity should be driven by error structure and governance needs, not by the number of available modules.

\subsection{Known Scenarios, Shared Projection, and Causal Boundaries}

The counterfactual evaluation of RQ3 is conducted in scenarios where the future path is known. If the future exogenous path is unknown, additional exogenous-prediction, data-assimilation, and uncertainty-propagation layers are needed. Shared projection is beneficial for comparing legally deployed outputs, but the candidate-state violation rate must be audited at the same time; otherwise a model that frequently produces illegal candidates may be ``fixed'' by projection without being noticed. Projection guarantees legality within the declared set and cannot discover omitted constraints.

The trackability of an intervention does not imply causal identifiability. Graph direction, confounder control, consistency, and intervenability require additional argument. This paper does not claim statistical identifiability, causal identification, universal approximation, or global stability. Operational identifiability only means reducing responsibility conflation through masking, freezing, ablation, and audit.

\subsection{Residual Governance and Ethics}

When the residual persistently hits the ceiling or stably clusters, candidate explanations---a missing rule, parameter drift, an observation fault, or a missing structural edge---should be proposed and promoted to a rule asset only after offline backtesting, constraint checking, and expert approval. The residual must not automatically modify the graph, automatically relax safety boundaries, or directly enter production deployment. Rule promotion should preserve the original evidence, version, approver, rollback path, and applicable domain.

The structural contract of this paper does not constitute an operational commitment to any real system; before real deployment, data quality, robustness, failure modes, responsibility allocation, and emergency fallback must still be verified.

\section{Conclusion}

This paper proposes an executable structural contract for graph world models, declaring node self-dynamics $S$, edge adjacency action $N$, and an optional bounded residual $R$ on a typed structural graph, synthesizing them via the composition operator $\mathcal{C}$, advancing through the unified one-step transition $T_{\Delta t}$, and converging via the global projection $\Pi_\Omega$, connecting rules, graph messages, domain solvers, and the learning residual into an auditable operational closed loop. The formal analysis gives a computational embedding with clear boundaries, feasible-set invariance, conditional error recursion, and a residual-amplitude bound. The frozen experiments produce positive evidence: compilation semantic fidelity, correct heterogeneous composition, end-to-end constraint and intervention consistency, structural fault localization outperforming baselines, the constrained residual reducing error without breaking constraints when the backbone has systematic bias, the declarative fixed-mechanism contract calibrated from observations on a semi-synthetic flood testbed having prediction accuracy no worse than free neural baselines with zero constraint violation, further validated on USGS public measured streamflow for multi-step prediction no worse than baselines with zero non-negativity violation, and validated for cross-conservation-law generalization on the real/verified topologies of the BattLeDIM water distribution network and the ACTIVSg2000 power transmission network, two different conservation laws---the same declared operator family is no worse than baselines in multi-step prediction under both mass and charge conservation with each $\Omega$ violation zero. It must be emphasized that this accuracy comparison is against baselines such as persistence, linear, MLP, and GCN\cite{ref6}, LSTM\cite{ref18}, STGCN\cite{ref17}, T-G-CN\cite{ref19} in small-sample semi-synthetic, single-real-watershed, or subgraph settings, and the value of the structural prior is not extrapolated to a universal accuracy advantage under big data or cross-domain. A key counter-evidence is that after expanding the USGS training period by about $14\times$ (18 to 254 days, the 254 days containing an extreme-flood segment, a distribution shift), the gap between the declarative backbone and the baselines widens rather than narrows, indicating that the structural prior is more robust under distribution shift---but it must be honestly noted that this comparison confounds data volume and distribution, and is not equivalent to a controlled verification under the same distribution that ``the advantage does not stem from small samples.'' Two honest secondary findings---the bounded and unbounded residuals performing comparably at this bias magnitude, and the local residual being weaker than the global---are also reported as is. Thus, what this paper advocates is a compilable, constrainable, and auditable structural contract, not a universal accuracy advantage. Future work should examine external validity on more conservation laws and real-city operational data, study the automatic selection of residual budget and scope, and establish tighter local error upper bounds. A model allowed to be everything cannot explain anything; auditability is bought at the price of declaring in advance what remains invariant.

\appendix
\renewcommand{\thesection}{\Alph{section}}

\section*{Appendix}
\addcontentsline{toc}{section}{Appendix}

\section{Reproduction and Audit}
\label{app:repro}

The results are jointly constrained by the frozen configurations, raw artifacts, statistical summaries, plotting scripts, and \texttt{results/manifest.json} (all under the companion code repository, not bundled in this arXiv source package). Each artifact records the relative path, SHA-256, generation command, configuration/source hash, UTC time, and evidence type; its timestamp and configuration hash allow a third party to verify the consistency of the two. It should be noted that this paper did not archive the configuration hash with a trusted third party before viewing the results (e.g., OSF or Zenodo), so ``the configuration was frozen before viewing the test results'' is currently self-attested by the authors---sufficient to constrain post-hoc cherry-picking, but not to rule out deliberate fabrication. The values of all result tables come from the frozen summary artifacts covered by the manifest: the statistical summary of RQ1--RQ5 (\texttt{results/summaries/rq1-rq5-summary.csv}, 113 rows, full mode) and the event metrics and summary JSONs of RQ6/RQ7 (\texttt{results/raw/} for each experiment family); the schematic and result figures use stable relative paths. When reproducing the statistics, the seeds or deterministic trajectories must be retained, and repetitions within a regime, method, or fault instance must not be counted as new independent samples.

\subsection{Hyperparameters}
\label{app:hyperparameters}

This appendix lists the hyperparameters of each research question, all fixed as code constants and not adjusted at experiment runtime. RQ1--RQ3 are deterministic synthetic or analytical experiments with no learnable parameters; RQ4 is deterministic diagnosis, with fault severity sampled by seed but the diagnosis method itself containing no learnable parameters; RQ5 trains the residual network.

\textbf{RQ5 residual-network training configuration.} See Table~\ref{tab:training-config}. The residual network is trained in float64 precision, single-threaded, with a deterministic algorithm; the optimizer is Adam and the loss is mean squared error. The global residual reads the whole graph (\texttt{GlobalResidualReader}, hidden-layer width 6), and the local residual is per-node independent (\texttt{LocalResidualReader}, hidden-layer width 5), with the same number of trainable parameters.

\begin{table}[h]
\centering
\caption{RQ5 residual-network training hyperparameters.}
\label{tab:training-config}
\begin{tabular}{lc}
\toprule
Hyperparameter & Value \\
\midrule
Hidden-layer width & 6 (global) / 5 (local) \\
Learning rate & $2\times10^{-2}$ \\
Training epochs (full) & 80 \\
Training samples (full) & 256 \\
Optimizer & Adam \\
Numerical precision & float64 \\
Residual budget $\varepsilon$ & 0.05 \\
\bottomrule
\end{tabular}
\end{table}

\textbf{Experiment grid and seeds.} See Table~\ref{tab:exp-grid}. All experiments share five model seeds $(11, 23, 37, 53, 71)$; RQ1--RQ3 are deterministic evidence, a single run agreeing bit-for-bit twice; the fault severity of RQ4 is sampled by seed, and the residual training and rollout of RQ5 use fixed seeds.

\begin{table}[h]
\centering
\caption{Grid size and seeds of each research question.}
\label{tab:exp-grid}
\begin{tabularx}{\linewidth}{@{}l l >{\raggedright\arraybackslash}X >{\raggedright\arraybackslash}X@{}}
\toprule
Research question & Seeds & Grid dimension & Statistical unit \\
\midrule
RQ1 & --- (deterministic) & 3 source models $\times$ multi-step trajectory & Deterministic state coordinate \\
RQ2 & --- (deterministic) & Heterogeneous graph $\times$ 12 transition steps & Deterministic transition step \\
RQ3 & --- (deterministic) & 4 projection strategies $\times$ 4 intervention scenarios & Intervention scenario \\
RQ4 & 5 & 6 fault classes $\times$ 3 methods & Seed-level diagnosis aggregation \\
RQ5 & 5 & 4 variants $\times$ 4 regimes $\times$ 8 events $\times$ 25 steps & Seed--regime rollout \\
\bottomrule
\end{tabularx}
\end{table}

\textbf{RQ5 crippled backbone and regime parameters.} The backbone spring-force edge weight is declared as $-1.5$ (truth $-2.0$, $25\%$ systematic bias); the damping coefficient is $0.1$, the cubic nonlinearity coefficient $\alpha$ varies by regime (iid $0.4$, missing-physics $0.8$, forcing-shift $0.4$, topology-shift $0.4$), the spring coefficient varies by regime (iid/missing-physics/forcing-shift $2.0$, topology-shift $2.6$), and the forcing amplitude varies by regime (iid/missing-physics/topology-shift $0.0$, forcing-shift $0.3$). The state legal domain is $[-3, 3]$, and the step size $\Delta t=0.05$.

\textbf{Statistical parameters.} The seed-level aggregation of RQ4 and RQ5 is the mean of five seeds; the fault-severity sampling of RQ4 uses \texttt{np.random.default\_rng(seed)}, and the training and rollout of RQ5 use \texttt{torch.manual\_seed(seed)} and a fixed NumPy seed, ensuring reproducibility.

\textbf{RQ6/RQ7 baseline and residual training configuration.} The neural baselines of RQ6 and RQ7 (MLP/GCN/LSTM/STGCN/T-G-CN) share the same training factory: hidden-layer width 32, Adam optimizer, learning rate $1\times10^{-3}$, at most 120 epochs, early-stopping patience 20, repeated on five seeds $(11,23,37,53,71)$, averaged within a seed over events/blocks first before entering the cross-event/block mean. The continuous backbone-parameter calibration of SD-GWM and the baseline training are separated into two non-joint paths: RQ6 calibrates from observation transitions by finite-difference least squares, and RQ7 fits the continuous parameters (per-station decay rate and baseflow bias, per-edge propagation weight) from training-period observation transitions by L-BFGS-B, with the integer routing delay locked independently per edge by a grid search over the single-step mean squared error on the training period on candidates $\{1,2,4,8,16,32,64\}$ steps (not entering the continuous optimization). The bounded residual $R$ training configuration of the two experiments is shown in Table~\ref{tab:rq67-config}.

\begin{table}[h]
\centering
\caption{RQ6/RQ7 bounded residual $R$ and evidence configuration.}
\label{tab:rq67-config}
\begin{tabularx}{\linewidth}{@{}l X X@{}}
\toprule
Configuration item & RQ6 (semi-synthetic flood) & RQ7 (USGS real data) \\
\midrule
Residual budget $\varepsilon$ & $5\%$ of node capacity & $10\%$ of that station's training-period std \\
Residual training optimizer/learning rate & Adam / $2\times10^{-2}$ & Adam / $2\times10^{-2}$ \\
Residual reader & Global, hidden 8 & Global, hidden 8 \\
Continuous-parameter calibration & Finite-difference least squares & L-BFGS-B \\
Integer delay determination & Fixed declaration & Grid search $\{1,2,4,8,16,32,64\}$ \\
Test statistical unit & 12 test events & 12 test time blocks \\
Evidence type & Semi-synthetic & Real \\
\bottomrule
\end{tabularx}
\end{table}

\section{Semi-Synthetic Flood Case Details}
\label{app:flood}

RQ6 elevates the flood case to the main evidence chain; this appendix supplements the details of its testbed and declarative contract. The $S$ and $N$ of this case are implemented as declarative fixed-mechanism operators under this paper's structural contract---runoff (\texttt{pipe\_self}: rainfall inflow plus staged drainage plus pump drainage), backwater and surface drainage (\texttt{ground\_self}), fill-difference gated conservative flow (\texttt{fill\_gated\_flow}), and threshold overflow (\texttt{threshold\_overflow})---rather than freely trained neural networks; only the authorized continuous parameters of these operators (admittance, drainage rates, thresholds, pump capacity, runoff coefficients, etc.) are calibrated from observation transitions, and the topology, operator types, and delays are fixed declarations.

\subsection{Testbed and Task Boundary}

\begin{figure}[h]
\centering
\includegraphics[width=0.95\linewidth]{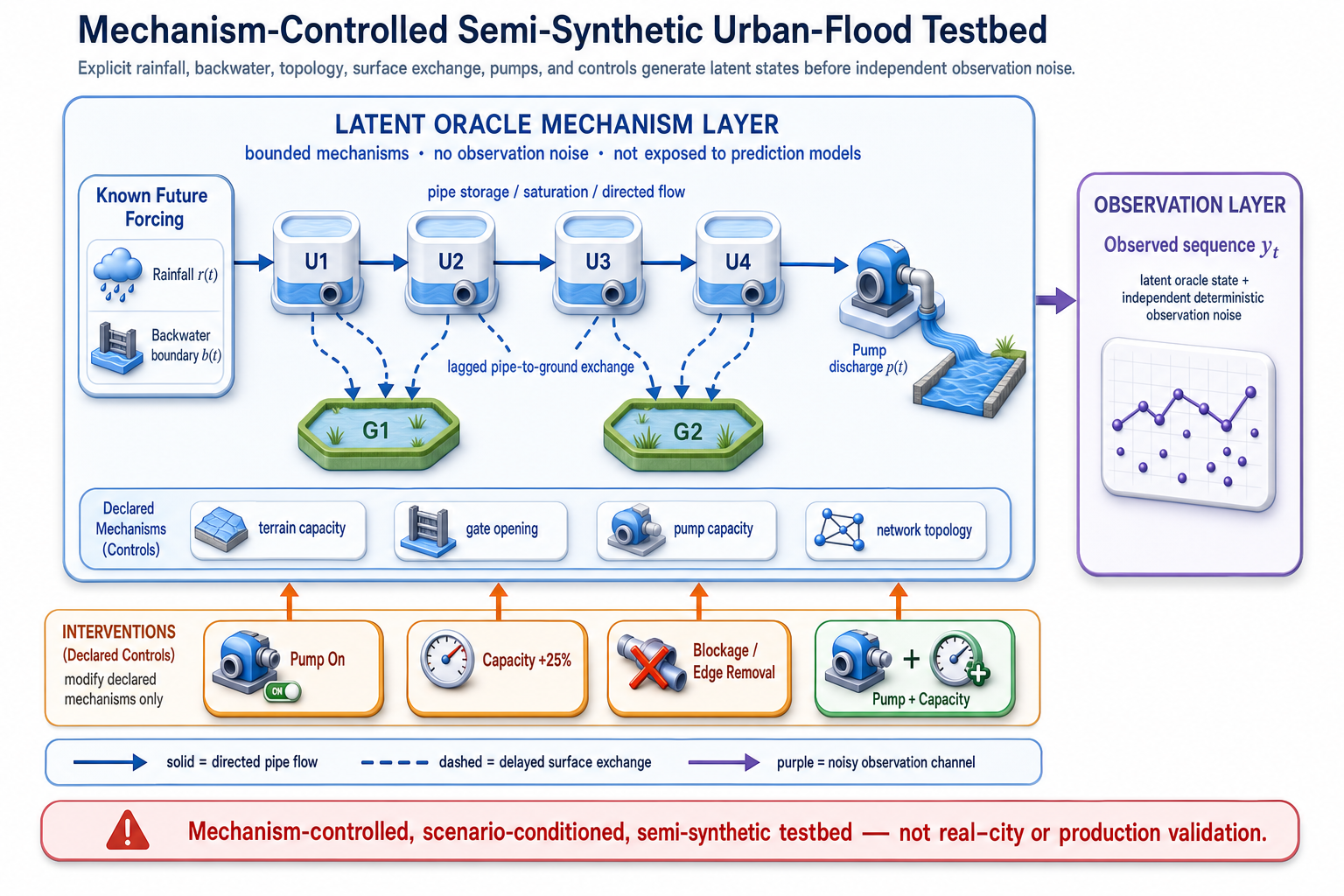}
\caption{Semi-synthetic flood testbed. The latent oracle state and the noisy observation are separated; rainfall, backwater, pump, and topological interventions are all explicitly declared.}
\label{fig:flood-testbed}
\end{figure}

The flood testbed consists of latent hydrodynamic states, explicit rainfall--pipe-network--surface-ponding--pump-station--river-channel mechanisms, and independent deterministic observation noise. It is a mechanism-controlled semi-synthetic testbed, not a real-city validation, and is not used for production deployment. The testbed generates sixty events (ten for each of six regime classes), split into 36/12/12 training, validation, and test sets by complete events, with the split done before windowing to prevent event leakage. Ten models are compared---persistence, linear, MLP, GCN\cite{ref6}, LSTM\cite{ref18}, STGCN\cite{ref17}, T-G-CN\cite{ref19}, and three configurations of this paper's method SD-GWM (w/o R, w/ R bounded residual $\varepsilon=0.05$, faithful backbone)---evaluated at 5, 15, and 30 minute prediction horizons; oracle physical roll-forward serves as a learning-free mechanism reference and does not enter the paired-test family. All models use the same input window, future driving path, and physical-clipping adapter. The testbed structure is shown in Figure~\ref{fig:flood-testbed}.

The task is scenario-conditioned forecasting: the future rainfall, river backwater, and control path are known in each scenario, and the conditional trajectories are then compared. It is not nowcasting in operational use, nor does it handle the real-time data-assimilation problem when the future rainfall path is unknown. The deployment metrics come from the physical-legalization adapter shared by all models; the candidate-state violation rate is recorded separately so that the shared projection does not hide model differences.

\section{Applicability Conditions of the Formal Statements}

Proposition~1 relies on fully matching the solver, step size, initial value, and input; Proposition~2 relies on the projection being correctly implemented and the constraint set being sufficiently declared; Proposition~3 holds only in the specified trajectory neighborhood and under local constants; Proposition~4 only bounds the residual contribution amplitude. Any use beyond these conditions should return to empirical verification rather than enlarging the theorem wording.

\section{Terminology and Cross-Language Freezing}

This English version shares all formulas, figures, tables, result numbers, and reference numbering with an internal Chinese draft; terminology follows the authoritative glossary in \texttt{terminology.csv} (25 canonical zh$\to$en pairs, fixing self-dynamics, neighbor graph-coupled dynamics, bounded residual correction, constrained rollout, feasible state space, semi-synthetic testbed, scenario-conditioned forecasting, operational identifiability, and other key renderings). No number, equation, table cell, or reference was changed in translation; the English version rewrites the syntax as academic English but does not alter the strength of any limitation or any negative result.

\end{document}